\documentclass{article} 
\usepackage[accepted]{icml2023}

\usepackage{amsmath,amsfonts,bm}

\def\eqref#1{equation~\ref{#1}}

\def\1{\bm{1}}

\def\vone{{\bm{1}}}

\DeclareMathAlphabet{\mathsfit}{\encodingdefault}{\sfdefault}{m}{sl}
\SetMathAlphabet{\mathsfit}{bold}{\encodingdefault}{\sfdefault}{bx}{n}

\newcommand{\E}{\mathbb{E}}

\DeclareMathOperator*{\argmax}{arg\,max}

\usepackage{hyperref}
\usepackage{url}

\usepackage{booktabs}       
\usepackage{amsfonts}       
\usepackage{nicefrac}       
\usepackage{microtype}      %
\usepackage{xcolor}         
\usepackage{tabularx}
\usepackage{graphicx}
\usepackage{subfigure}
\usepackage{mathrsfs}
\usepackage{amsmath,amssymb,amsthm,amsbsy,latexsym,dsfont,tikz}
\usepackage{wrapfig,lipsum}

\usepackage{enumerate}

\usepackage{enumitem}

\newcommand{\diag}[0]{\text{diag}}
\newcommand{\Id}[0]{\text{Id}}

\newcommand{\rank}[0]{\text{rank}}

\theoremstyle{remark}
\newtheorem{thm}{Theorem}
\newtheorem{lem}[thm]{Lemma}

\newtheorem{prop}[thm]{Proposition}

\newtheorem{rem}[thm]{Remark}

\renewcommand{\leq}{\leqslant} 
\renewcommand{\geq}{\geqslant}

\def\qed{ \hfill $\blacksquare$}  

\newcommand{\vX}{\mathbf{X}}\newcommand{\vY}{\mathbf{Y}}

\newcommand{\bP}{\mathbb{P}}\newcommand{\bR}{\mathbb{R}}

\DeclareMathOperator{\tr}{tr} 

\icmltitlerunning{Likelihood Adjusted SDP for Clustering Heterogeneous Data}

\begin{document}

\twocolumn[
\icmltitle{Likelihood Adjusted Semidefinite Programs for Clustering Heterogeneous Data}




\begin{icmlauthorlist}
\icmlauthor{Yubo Zhuang}{yyy}
\icmlauthor{Xiaohui Chen}{yyy,xxx}
\icmlauthor{Yun Yang}{yyy}
\end{icmlauthorlist}
\icmlaffiliation{xxx}{Department of Mathematics, University of Southern California}
\icmlaffiliation{yyy}{Department of Statistics, University of Illinois at Urbana-Champaign.}

\icmlcorrespondingauthor{Yubo Zhuang}{yubo2@illinois.edu}
\icmlcorrespondingauthor{Xiaohui Chen}{xiaohuic@usc.edu}
\icmlcorrespondingauthor{Yun Yang}{yy84@illinois.edu}
\icmlkeywords{clustering, likelihood inference, semidefinite programming, alternating maximization}

\vskip 0.3in
]



\printAffiliationsAndNotice{}  

\begin{abstract}
Clustering is a widely deployed unsupervised learning tool. Model-based clustering is a flexible framework to tackle data heterogeneity when the clusters have different shapes. Likelihood-based inference for mixture distributions often involves non-convex and high-dimensional objective functions, imposing difficult computational and statistical challenges. The classic expectation-maximization (EM) algorithm is a computationally thrifty iterative method that maximizes a surrogate function minorizing the log-likelihood of observed data in each iteration, which however suffers from bad local maxima even in the special case of the standard Gaussian mixture model with common isotropic covariance matrices. On the other hand, recent studies reveal that the unique global solution of a semidefinite programming (SDP) relaxed $K$-means achieves the information-theoretically sharp threshold for perfectly recovering the cluster labels under the standard Gaussian mixture model. In this paper, we extend the SDP approach to a general setting by integrating cluster labels as model parameters and propose an iterative likelihood adjusted SDP (iLA-SDP) method that directly maximizes the \emph{exact} observed likelihood in the presence of data heterogeneity. By lifting the cluster assignment to group-specific membership matrices, iLA-SDP avoids centroids estimation -- a key feature that allows exact recovery under well-separateness of centroids without being trapped by their adversarial configurations. Thus iLA-SDP is less sensitive than EM to initialization and more stable on high-dimensional data. Our numeric experiments demonstrate that iLA-SDP can achieve lower mis-clustering errors over several widely used clustering methods including $K$-means, SDP and EM algorithms.


\end{abstract}

\section{Introduction}
\label{sec:intro}

Clustering analysis has been widely studied and regularly used in machine learning and its applications in network science~\citep{doi:10.1073/pnas.122653799}, computer vision~\citep{868688,5539868}, manifold learning~\citep{CHEN2021303} and bioinformatics~\citep{10.1093/bib/bbz170}. Perhaps by far the most popular clustering method is the $K$-means~\citep{MacQueen1967_kmeans} partially because there are computationally convenient algorithms such as Lloyd's algorithm and $K$-means++ for heuristic approximation~\citep{Lloyd1982_TIT,10.5555/1283383.1283494}. Mathematically, $K$-means aims to find the optimal partition of data to minimize the total within-cluster squared Euclidean distances, which is equivalent to the maximum profile likelihood estimator under the standard Gaussian mixture model (GMM) with common isotropic covariance matrices~\citep{chen2021cutoff}. Nevertheless, real data usually exhibit various degrees of heterogeneous features such as the cluster shapes may vary from component to component, which renders $K$-means as a sub-optimal clustering method.

Another popular clustering method is the classic expectation-maximization (EM) algorithm, which is a computationally thrifty method based on the idea of data augmentation to iteratively optimize the non-convex observed data likelihood~\citep{EM1977}. Theoretical investigations reveal that the EM algorithm suffers from bad local maxima even in the one-dimensional standard GMM with well-separated cluster centers~\citep{JinZhangSilvaramanWainwrightJordan2016_EM}. Thus practically even when applied in highly favorable separation-to-noise ratio settings, careful initialization, often through multiple random initializations or a warm-start by another heuristic method such as hierarchical clustering~\citep{FraleyRaftery2002}, is the key for the EM algorithm to find the correct cluster labels and model parameters. With a reasonable initial start, the EM algorithm has been shown to achieve good statistical properties~\citep{10.1214/16-AOS1435,https://doi.org/10.48550/arxiv.1908.10935}.

In this paper, we consider the likelihood-based inference to tackle the problem of recovering cluster labels in the presence of data heterogeneity. Our motivation stems from the recent progress in understanding the computational and statistical limits for convex relaxation methods of the $K$-means clustering. Since $K$-means is a worst-case NP-hard problem~\citep{aloise2009np}, various heuristic approximation algorithms such as Lloyd’s algorithm~\citep{Lloyd1982_TIT,LuZhou2016}, and computationally tractable relaxations such as spectral clustering~\citep{Meila01learningsegmentation,NgJordanWeiss2001_NIPS,Vempala04aspectral,Achlioptas2005McSherry,vanLuxburg2007_spectralclustering,vonLuxburgBelkinBousquet2008_AoS} and semidefinite programs (SDP)~\citep{PengWei2007_SIAMJOPTIM,MixonVillarWard2016,LiLiLingStohmerWei2017,FeiChen2018,CHEN2021303,Royer2017_NIPS,GiraudVerzelen2018,BuneaGiraudRoyerVerzelen2016,ZhuangChenYang2022}, have been proposed in literature. Among the existing solutions, the SDP approach is particularly attractive in that it attains information-theoretically optimal threshold on centroid separations for exact recovery of cluster labels~\citep{chen2021cutoff}.

{\bf Our contributions.} We extend the SDP approach to a general setting with heterogeneous features by integrating cluster labels as model parameters (together with other component-specific parameters) and propose an \emph{iterative likelihood adjusted SDP} (iLA-SDP) method that directly maximizes the \emph{exact} observed data likelihood. Our idea is to tailor the strength of SDP relaxation of the $K$-means clustering method in the isotropic covariance case for likelihood-awareness inference. On one hand, iLA-SDP has a similar flavor as the EM algorithm by maximizing the likelihood function of the observed data. On the other hand, different from the EM framework, iLA-SDP treats the cluster labels as \emph{unknown} parameters while profiles out the cluster centers (i.e., centroids), which brings several statistical and algorithmic advantages. 

First in the arguably simplest one-dimensional GMM setting, EM is known to fail in certain configurations of centroids even when they are well-separated~\citep{JinZhangSilvaramanWainwrightJordan2016_EM}. In other words, EM is sensitive to initialization and model configuration. The main reason is due to the effort for estimating the cluster centers during the EM iterations. In iLA-SDP, cluster centers are regarded as nuisance parameters and profiled out to obtain a likelihood function in component-specific parameters including only the cluster covariance matrices. 
Thus iLA-SDP is more stable and performs empirically better than EM.

Second, cluster labels in EM are latent variables that are estimated by their posterior probabilities and the observed log-likelihood for component parameters and mixing weights are optimized through minorizing functions during iterations. In iLA-SDP, cluster labels are regarded as parameters optimized through the likelihood function jointly in the labels and covariance matrices. Thus iLA-SDP is a more direct approach than EM for taming the non-convexity in the observed log-likelihood objective and we prove that it perfectly recovers the true clustering structure if the clusters are well-separated under a lower bound without concerning the configurations of centroids.

The rest of the paper is organized as follows. In Section~\ref{sec:likelihood_model}, we review some background on partition-based formulation for model-based clustering. In Section~\ref{sec:LA-SDP}, we introduce the likelihood adjusted SDP for recovering the true partition structure and discuss its connection to the EM algorithm. 
In Section~\ref{sec:real_data}, we compare the performance of several widely used clustering methods on two real datasets.

\section{Model-based clustering: a partition formulation}
\label{sec:likelihood_model}

We consider the model-based clustering problem. Suppose the data points $X_1, \dots, X_n \in \mathbb{R}^p$ are independent random variables sampled from $K$-component Gaussian mixture model (GMM). Specifically, let $G_1^*, \dots, G_K^*$ be the true partition of the index set $[n] := \{1, \dots, n\}$ such that if $i \in G_k^*$, then
\begin{equation}
    \label{eqn:gmm}
    X_i=\mu_k+\epsilon_i,
\end{equation}
where $\mu_k \in \bR^p$ is the center of the $k$-th cluster and $\epsilon_i$ is an i.i.d. random noise term following the common distribution $N(0,\Sigma_k)$. Here we focus on the most general and realistic scenario where the within-cluster covariance matrices $\Sigma_1, \dots, \Sigma_K$ are heterogeneous. In our formulation of the GMM, the true partition $(G_k^*)_{k=1}^K$ is treated as a \emph{unknown} parameter in model~(\ref{eqn:gmm}), along with the component-wise parameters $(\mu_k, \Sigma_k)_{k=1}^K$. With this parameterization $(G_k, \mu_k, \Sigma_k)_{k=1}^K$, the log-likelihood function for observing the data $\vX = \{X_1, \dots, X_n\}$ is given by 

$\ell\big((G_k, \mu_k, \Sigma_k)_{k=1}^K \mid \vX \big) =  -\sum_{k=1}^{K}\frac{|G_k|}{2}\log(2\pi |\Sigma_k|)
-\frac{1}{2} \sum_{k=1}^K \sum_{i \in G_k} (X_i-\mu_k )^T \Sigma_k^{-1} (X_i-\mu_k),$

where $|G_k|$ is the cardinality of $G_k$ and $|\Sigma_k|$ is the determinant of matrix $\Sigma_k$. Since we are primarily interested in recovering the clustering labels (or equivalently the assignment matrix, cf. Section~\ref{subsec:LA-SDP_oracle} below) in the presence of cluster heterogeneity, we can first profile out the nuisance parameters $\mu_k$ in closed form and the resulting objective function as a profile log-likelihood for the remaining parameters (after dropping constants) is given by
\begin{align*}
&\ell\big((G_k, \Sigma_k)_{k=1}^K \mid \vX \big) = -\sum_{k=1}^K|G_{k}|\log(|\Sigma_k| )\\
&-\sum_{k=1}^K\sum_{i\in G_k}\| X_i\|^2_{\Sigma_k^{-1}}+\sum_{k=1}^K\frac{1}{|G_k|}\sum_{i,j\in G_k}\langle X_i,X_j \rangle_{\Sigma_k^{-1}},  \stepcounter{equation}\tag{\theequation}\label{eqn:adj_kmeans}
\end{align*}
where $\langle v,u \rangle_{\Sigma}:=v^T \Sigma u$ and $\|u\|^2_{\Sigma}:=\langle u,u \rangle_{\Sigma}$ for any $u,v\in \mathbb{R}^p$ and $\Sigma\succ 0$. This leads us to a combinatorial optimization problem for the profile log-likelihood function of the following form:
\begin{equation}
    \label{eqn:profile_loglik}
    \max \left\{ \ell\big ((G_k, \Sigma_k)_{k=1}^K \mid \vX \big) : \bigsqcup_{k=1}^{K} G_{k} = [n], \; \Sigma_k \succ 0 \right\},
\end{equation}
where the disjoint union $\bigsqcup_{k=1}^{K} G_{k} = [n]$ means that $\bigcup_{k=1}^{K} G_{k} = [n]$ and $G_j \cap G_k = \emptyset$ if $j \neq k$. Note that the constrained optimization problem in~(\ref{eqn:profile_loglik}) in the special case $\Sigma_1 = \cdots = \Sigma_k = \sigma^2 \Id_p$ reduces to the $K$-means clustering method, which is known to be worst-case NP-hard~\citep{Dasgupta2007,MahajanNimbhorkarVaradarajan2009}. To overcome such computational difficulty, semidefinite program (SDP) relaxation is a tractable solution that achieves information-theoretically optimal exact recovery under the standard GMM with identical and isotropic covariance matrices~\citep{CHEN2021303}. Nevertheless, all existing formulations of various SDP relaxations of the standard GMM critically depend on the assumption that $\Sigma_1 = \cdots = \Sigma_k = \sigma^2 \Id_p$ with a \emph{known} noise variance parameter $\sigma^2$~\citep{FeiChen2018,LiLiLingStohmerWei2017,PengWei2007_SIAMJOPTIM,CHEN2021303}. This motivates us to seek alternative SDP formulations adjusting the (full) information coming from the likelihood function for the observed data $\vX$.

\section{Likelihood adjusted SDP for clustering heterogeneous data}
\label{sec:LA-SDP}

In this section, we introduce the likelihood adjusted SDP (LA-SDP) for recovering the true partition structure $G_1^*, \dots, G_K^*$ by applying convex relaxation to the profile log-likelihood function~(\ref{eqn:profile_loglik}).

\subsection{Oracle LA-SDP under known covariance matrices}
\label{subsec:LA-SDP_oracle}
In this subsection, we consider the oracle case where the covariance matrices $\Sigma_1, \dots, \Sigma_K$ are known. 
Let us start with a well-studied SDP relaxation formulation~\citep{PengWei2007_SIAMJOPTIM} for approximating the combinatorial optimization problem of maximizing the profile log-likelihood function under the isotropic setting with known $\Sigma_1=\ldots=\Sigma_K=\sigma^2 \Id_p$, which is known~\citep{chen2021cutoff} to attain the information-theoretically optimal threshold on centroid separations for exact recovery of cluster labels. Note that there is a one-to-one correspondence between any given partition $(G_k)_{k=1}^K$ of $[n]$ and a binary assignment matrix $H=(h_{ik})\in \{0,1\}^{n\times K}$ (up to cluster labels permutation) such that $h_{ik} = 1$ if $i \in G_k$ and $h_{ik} = 0$ otherwise for $i \in [n]$ and $k \in [K]$. Because each row of $H$ contains exactly one non-zero entry, the recovery of the true clustering structure (or its associated assignment matrix) by maximizing the profile log-likelihood function (after dropping constants) can be re-expressed as a (non-convex) mixed integer program:
\begin{equation*}
\begin{gathered}
\max_{H} \ \langle A, H B H^\top \rangle =  \sum_{k=1}^K\frac{1}{|G_k|}\sum_{i,j\in G_k}\langle X_i,X_j\rangle, \\
\mbox{subject to } H \in \{0,1\}^{n \times K} \mbox{ and } H \vone_{K} = \vone_{n}, 
\end{gathered}
\stepcounter{equation}\tag{\theequation}\label{eqn:kmeans-mixed_integer}
\end{equation*}
where $A = X^\top X$ is the $n \times n$ similarity matrix, $\vone_n$ denotes the $n$-dimensional vector of all ones, and $B$ is the $K\times K$ diagonal matrix whose $k$-th diagonal component is $|G_k|^{-1} = \big(\sum_{i=1}^nh_{ik}\big)^{-1}$. Here, we have used the key identity $\sum_{k=1}^K w_{k} \sum_{i,j \in G_k} a_{ij} = \langle A, H B H^\top \rangle$
that holds for any diagonal matrix $B=\diag(w_1,\ldots,w_K)$ and similarity matrix $A = (a_{ij})_{i,j=1}^n$.
Relaxing the above mixed integer program~(\ref{eqn:kmeans-mixed_integer}) by lifting the assignment matrix $H$ into $Z = H B H^\top$, we arrive at its SDP relaxation as
\begin{equation*}
\begin{gathered}
\hat{Z} = \arg\max_{Z \in \bR^{n \times n}} \langle A, Z \rangle, \\
\mbox{subject to } Z \succeq 0, \; \tr(Z) = K, \; Z \vone_{n} = \vone_{n}, \; Z \geq 0, 
\end{gathered}
\stepcounter{equation}\tag{\theequation}\label{eqn:kmeans_sdp}
\end{equation*}
where $Z \geq 0$ means each entry $Z_{ij} \geq 0$ and $Z \succeq 0$ means the matrix $Z$ is symmetric and positive semi-definite. This SDP formulation relaxes the integer constraint on $H$ into two linear constraints $\tr(Z) = K$ and $Z\geq 0$ that are satisfied by any $Z=HBH^T$ as $H$ ranges over feasible solutions of problem~(\ref{eqn:kmeans-mixed_integer}).

Now let us consider the general heterogeneous setting with (possibly) different and non-isotropic covariance matrices $\Sigma_1, \dots, \Sigma_K$, and extend the SDP relaxation to this setting. Two technical difficulties arise by examining the previous argument. First, the first two terms in the profile log-likelihood function~(\ref{eqn:profile_loglik}) are no longer independent of the assignment matrix, and is therefore not negligible. In particular, they also provide partial information about the cluster labels when the covariance matrices are different: $\| X_i\|^2_{\Sigma_k^{-1}}$ in the second term quantifies how well $X_i$ aligns with the covariance matrix $\Sigma_k$ encoding second-order information of the $k$-th cluster; while the first term plays the role of balancing the cluster sizes and favors assigning more points to clusters with smaller shapes (since density is expected to be high).
Second, the similarity $\langle X_i,X_j \rangle_{\Sigma_k^{-1}}$ within cluster $G_k$ in the third term now depends on $k$, making the key identity $\sum_{k=1}^K w_{k} \sum_{i,j \in G_k} a_{ij} = \langle A, H B H^\top \rangle$ for connecting the profile log-likelihood function with the objective function of the mixed integer program~(\ref{eqn:kmeans-mixed_integer}) no longer applicable.

To solve the two aforementioned difficulties, we propose to augment the single variable $Z$ in the SDP relaxation~(\ref{eqn:kmeans_sdp}) to $K$ variables $(Z_k)_{k=1}^K$, where $Z_k$ can be interpreted as the lifting of the $k$-th column $H_k$ of the assignment matrix $H$ via $Z_k = \frac{1}{|G_k|} H_k H_k^\top$, $|G_k|=\sum_{i=1}^n h_{ik} = H_k^\top\vone_{n}$, that encodes the cluster membership associated with the $k$-th cluster. More specifically, by extending the key identity in the isotropic setting to $\sum_{k=1}^K w_{k} \sum_{i,j \in G_k} a_{ij}^{(k)} =\sum_{k=1}^K \langle A^{(k)}, H_k w_k H_k^\top \rangle$ for any weight vector $w = (w_k)_{k=1}^K$ and $K$ similarity matrices $\big(A_k = (a^{(k)}_{ij})_{i,j=1}^n\big)_{k=1}^K$, we can analogously express the maximizing profile log-likelihood problem as the following (non-convex) mixed integer program:
\begin{equation*}
\begin{gathered}
\max_{H} \ \sum_{k=1}^K\langle A^{(k)}, H_k w_k H_k^\top \rangle, \\
\mbox{subject to } H_k \in \{0,1\}^{n \times 1} \mbox{ and } \sum_{k=1}^K H_k = \vone_{n},
\end{gathered}
\stepcounter{equation}\tag{\theequation}\label{eqn:kmeans-mixed_integer2}
\end{equation*}
where $w_k = |G_k|^{-1}=\big(\sum_{i=1}^n h_{ik}\big)^{-1}$, and the $k$-th cluster-specific similarity matrix $A^{(k)} :=$
\begin{equation}
\label{eqn:A_k}
 - \log(|\Sigma_k|)\vone_{n} \vone_{n}^T- \frac{1}{2}\left[v_k\vone_{n}^T+\vone_{n}v_k^T\right]+ X^T \Sigma_k^{-1} X,
\end{equation}
where $v_k:=\diag(X^T \Sigma_k^{-1} X)$, $\diag(A)$ stands for the column vector composed of all diagonal entries of a matrix $A$. Now by lifting $H_k$ into $Z_k = H_k w_k H_k^\top$, we arrive at the following SDP relaxation for the profile log-likelihood objective function~(\ref{eqn:adj_kmeans}):
\begin{equation*}
\begin{gathered}
\big(\hat{Z}_1,\dots,\hat{Z}_K\big) = \argmax_{Z_1,\dots Z_K \in \bR^{n \times n}} \sum_{k=1}^K\langle A_k, Z_k \rangle, \\
\mbox{subject to }  \sum_{k=1}^K\tr(Z_k) = K,\; \big(\sum_{k=1}^K Z_k\big) \vone_{n} = \vone_{n},\\
Z_k \succeq 0, \; Z_k \geq 0, \; \forall \; k\in [K],
\end{gathered}
\stepcounter{equation}\tag{\theequation}\label{eqn:kmeans_asdp}
\end{equation*}
which relaxes the integer constraint on $H=(H_1,H_2,\cdots,H_k)$ into $(K+1)$ linear constraints $\sum_{k=1}^K\tr(Z_k) = K$ and $Z_k\geq 0$ for $k\in[K]$ that are satisfied by any $Z_k=H_kw_kH_k^\top$ as $H$ ranges over feasible solutions of problem~(\ref{eqn:kmeans-mixed_integer2}).

Since solving~(\ref{eqn:kmeans_asdp}) requires the knowledge of the true covariance matrix for each component, we call the solution $(\hat{Z}_k)_{k=1}^K$ as the \emph{oracle} likelihood adjusted SDP (LA-SDP) for estimating the cluster membership matrix of data points. In the special case of isotropic covariance matrices $\Sigma_1 = \cdots = \Sigma_K = \sigma^2 \Id_p$, Proposition~\ref{prop:asdp_to_sdp} below shows that LA-SDP reduces to become equivalent to the previous SDP formulation~(\ref{eqn:kmeans_sdp}).

\begin{figure}[h!] 
   \vspace{0.1cm}
   \centering
   \includegraphics[trim={2cm 1.4cm 2cm 1.4cm},clip,scale=0.135]{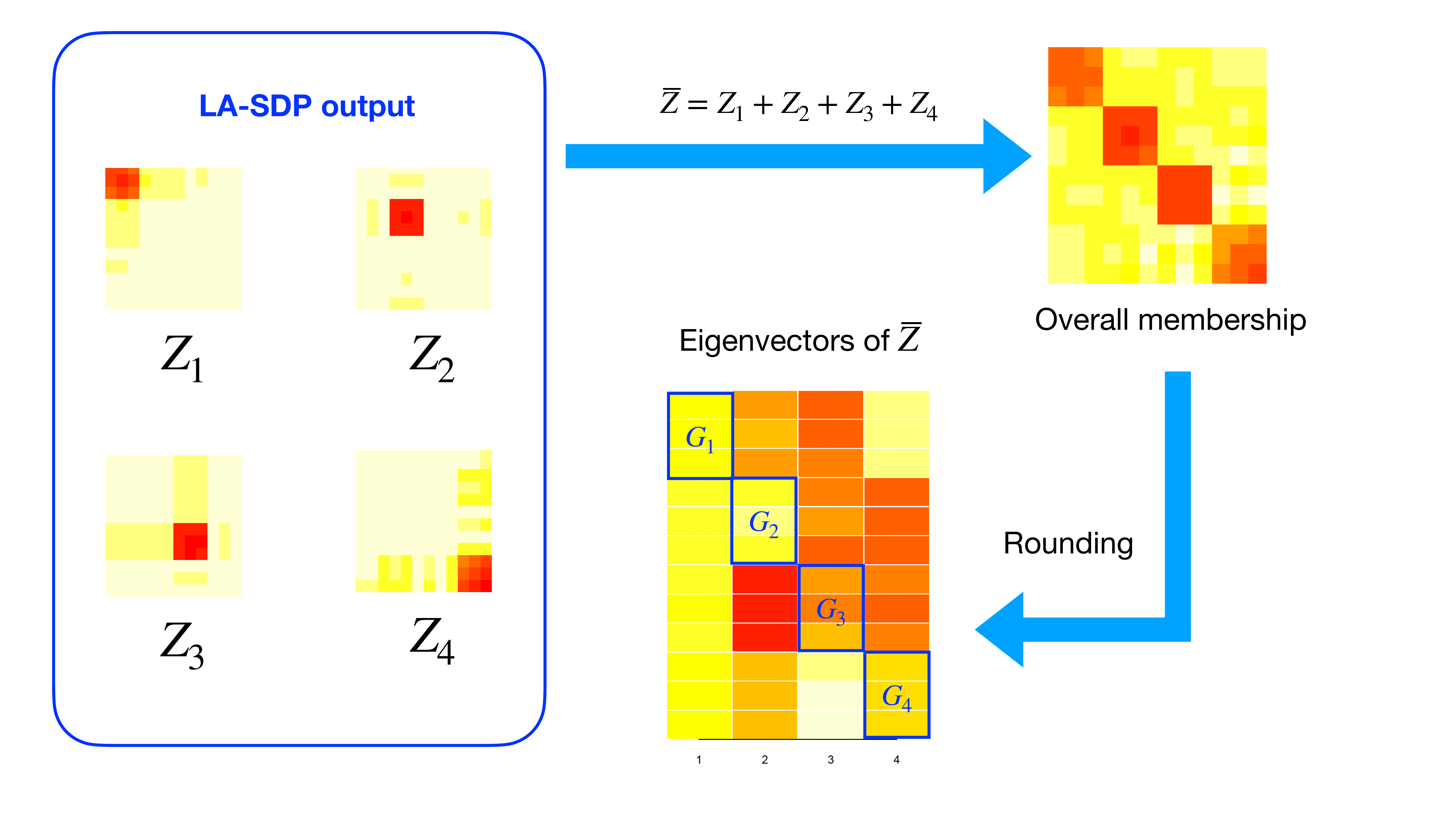}
   \caption{LA-SDP membership matrices to cluster labels via spectral rounding.}
   \label{fig:LA-SDP membership matrices}
   \vspace{0.1cm}
\end{figure}

\begin{prop}[\bf SDP relaxation for $K$-means is a special case of LA-SDP]\label{prop:asdp_to_sdp}
Suppose $\Sigma_k = \sigma^2 \Id_p$ for all $k\in[K]$. Let $\hat{Z}$ be the solution to~(\ref{eqn:kmeans_sdp}) that achieves maximum $M_1$ and $\hat{Z}_k,k=1,\dots,K,$ be the solution to~(\ref{eqn:kmeans_sdp}) with maximum $M_2$. Then $M_1=M_2$. And $\hat{Z}=\sum_{k=1}^K \hat{Z}_k$, if $\hat{Z}$ is unique in~(\ref{eqn:kmeans_sdp}).
\end{prop}

Note that the SDP relaxed $K$-means in~(\ref{eqn:kmeans_sdp}) is originally proposed in~\citep{PengWei2007_SIAMJOPTIM} and has been extensively studied in literature. In particular, it achieves the information-theoretical limit for exact recovery under the standard GMM~\citep{CHEN2021303} and it is robust against outliers and adversarial attack~\citep{FeiChen2018}. In the case of exact recovery where $\hat{Z} = Z^*$ and $Z^*$ is the true cluster membership matrix such that $Z^*_{ij} = |G_k^*|^{-1}$ if $i, j \in G_k^*$ and $Z^*_{ij} = 0$ otherwise, then we can easily recover the true partition structure $G_1^*, \dots, G_K^*$ or its associated assignment matrix from the block diagonal matrix $\hat{Z}$. Thus it is an interesting theoretical question of when the partition structure induced by $\hat Z =\sum_{k=1}^K\hat Z_k$ from the LA-SDP (see Figure~\ref{fig:LA-SDP membership matrices} for an illustration) can achieve exact recovery. Theorem~\ref{thm:main_thm} below gives a lower bound of the separation signal-to-noise ratio for achieving exact recovery in the presence of data heterogeneity.

For each distinct pair $(k,l)\in [K]$, let $D_{(k,l)}:= \frac{\sum_{i=1}^p\left(\lambda_i -\log(1+\lambda_i)\right)}{p\max_{i}|\lambda_i |}$ characterize the closeness between $\Sigma_k$ and $\Sigma_l$, where $\lambda_1,\ldots,\lambda_p$ enumerate all eigenvalues of $(\Sigma_l^{1/2}\Sigma_k^{-1}\Sigma_l^{1/2}-\Id_p )$. If $\lambda_i=0,\;\forall i\in [p],$ we let $D_{(k,l)}=0$. Let $\Delta^2:= \min_{k\ne l}\|\Sigma_k^{-1/2}(\mu_k-\mu_l) \|^2$ denote a covariance adjusted centroid separation, $n_k:=|G_k^*|$ the size of true cluster $G_k^\ast$, $m=\min_{k\ne l}\frac{2n_kn_l}{n_k+n_l}$ the least pairwise harmonic mean over cluster sizes, $\underline{n}=\min_{k} n_k$ the minimal cluster size, and $M:= \max_{k \ne l}\|\Sigma_l^{1/2}\Sigma_k^{-1}\Sigma_l^{1/2} \|_{\rm op}$ (matrix operator norm).

\begin{figure}[h!] 
   \vspace{-0.1cm}
   \centering
      \subfigure{\includegraphics[trim={1.65cm 7.5cm 1.65cm 7cm},clip,scale=0.4]{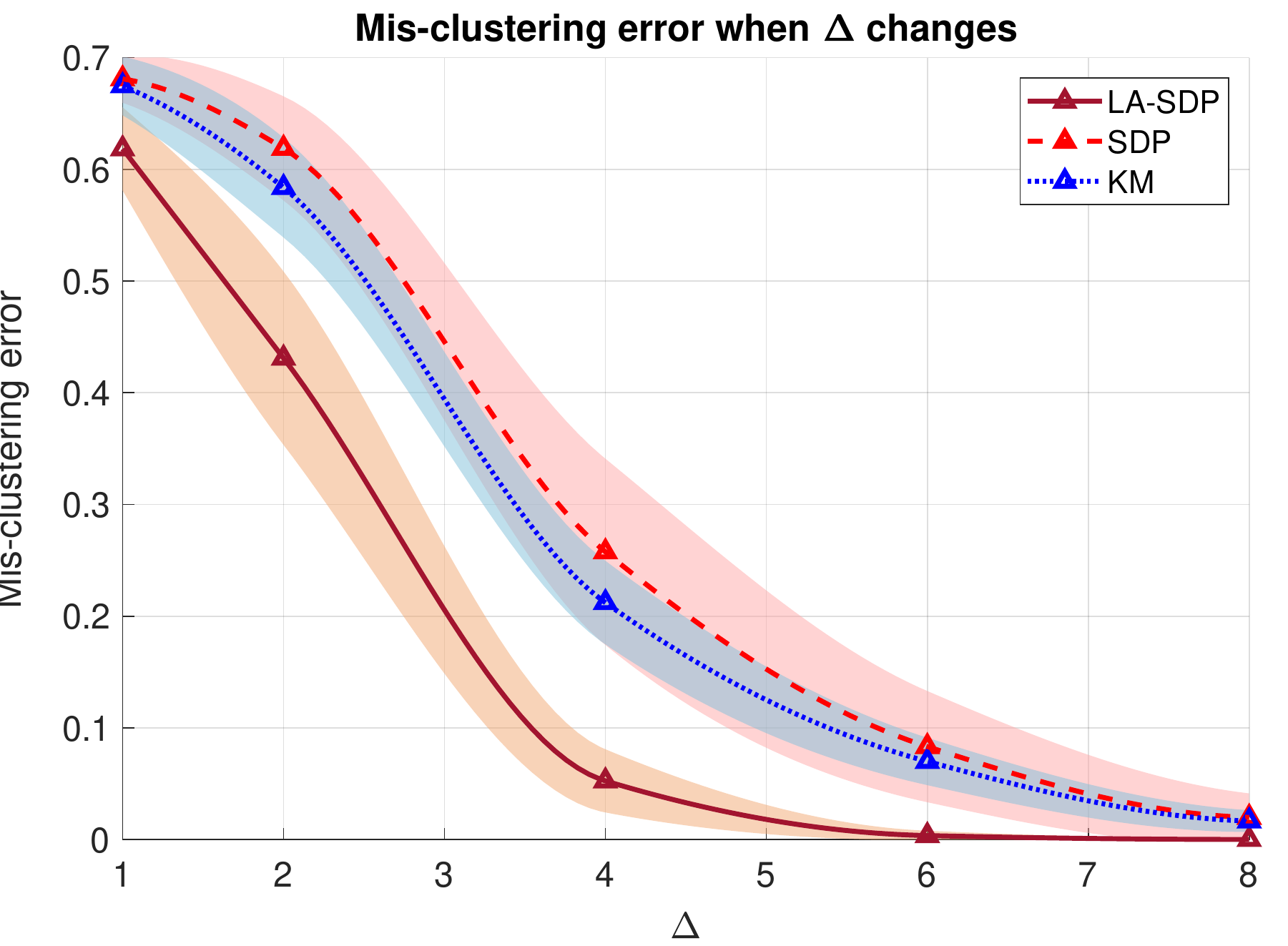}} 
      \subfigure{\includegraphics[trim={1.65cm 6cm 1.65cm 7cm},clip,scale=0.4]{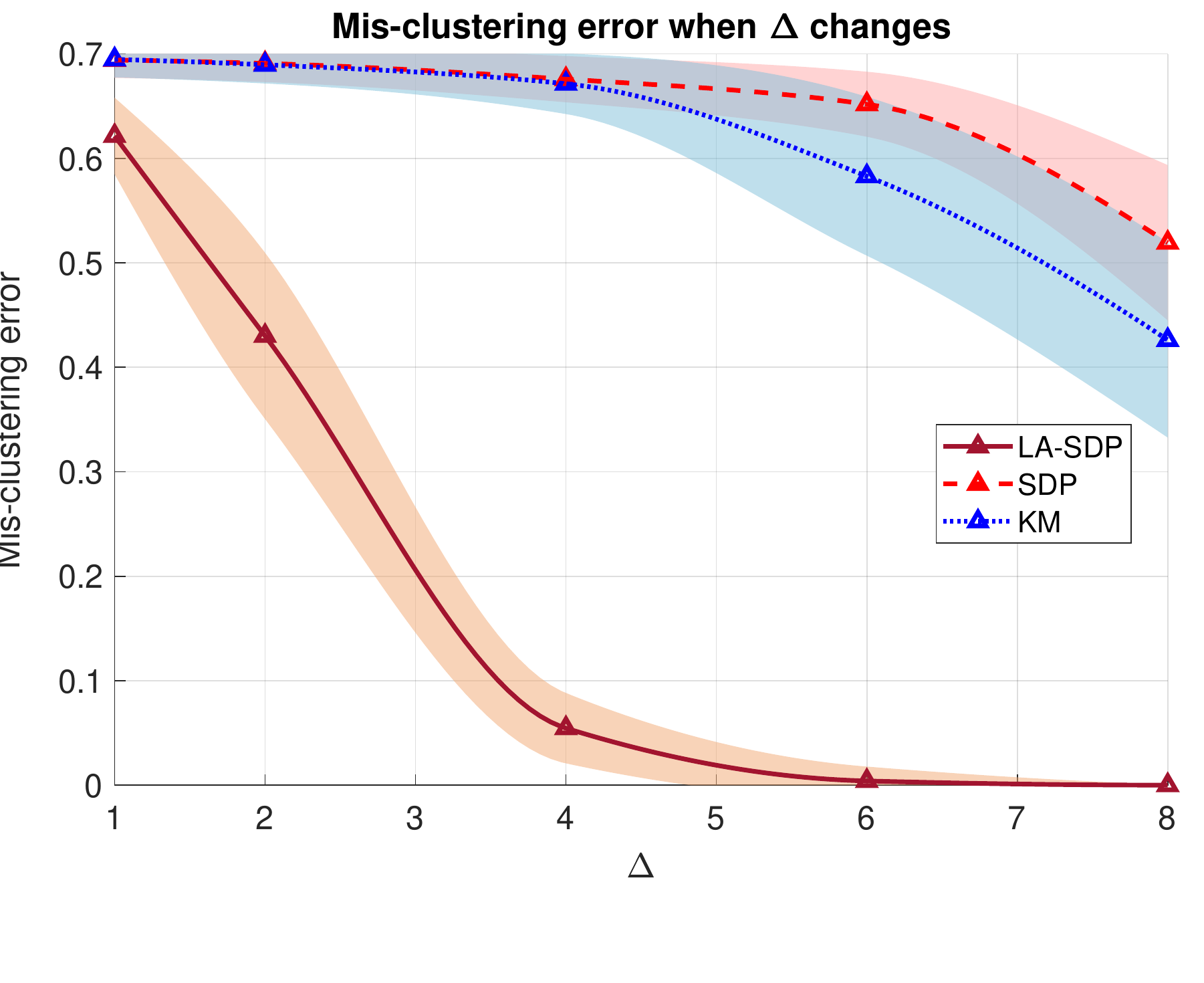}}\\[-7ex]
\caption{Mis-clustering error (with shaded error bars) vs centroid separation $\Delta$ under different conditional numbers of cluster covariance matrices $\Sigma_1=\Sigma_2=\cdots=\Sigma_K$ ($M=1$). The top (bottom) plot corresponds to a moderate (large) condition number of the common covariance matrix. Here, KM refers to $K$-means method; SDP refers to the original SDP~(\ref{eqn:kmeans_sdp}).}
   \label{fig:sig_to_noise1}
\end{figure}

\begin{thm}[\bf Exact recovery for LA-SDP]\label{thm:main_thm}
Suppose there exist constants $\delta >0$, $\beta\in(0,1)$ and $\eta\in(0,1)$ such that
\[
\log n \geq \max\left\{\frac{(1-\beta)^2}{\beta^2}, \frac{(1-\beta)(1-\eta)K^2}{\beta^2 \max\{(M-1)^2,1\}}  \right\}\frac{ C_1 n}{ m},
\]
\[
\delta\leq  \frac{\beta^2}{(1-\beta)^2}\frac{ C_2 M^{1/2}}{ K},\; m\geq\frac{4(1+\delta)^2}{\delta^2}.
\]
Then the LA-SDP achieves exact recovery, or $\hat Z=Z^*$, with probability at least $1-C_7 K^3 n^{-\delta}$ if
\begin{equation}
\begin{aligned}
\label{eqn:lower_bound_SNR}
\Delta^2\geq (E_1+E_2)\log n,\;\min_{k\ne l} D_{(k,l)}\geq C_3(1+\frac{\log n}{p}+\frac{p}{n}),
\end{aligned}
\end{equation}
where concrete expressions of $E_1$ and $E_2$ (depending on $\delta,\beta,\eta$) are provided in Appendix~\ref{appendix:proof}, and $C_1,\dots,C_7$ are universal constants. 
\end{thm}
In general, the upper bound condition on $n/m$ requires the cluster sizes tend to be balanced in which case $n/m =K$. Our definition of the centroid separation $\Delta$ extends the separation-to-noise ratio (SNR) for the exact recovery under the isotropic covariance setting~\citep{CHEN2021303} to the heterogeneous setting by taking into account the cluster shapes (i.e.~second order information).  From~(\ref{eqn:lower_bound_SNR}), we see that our theoretical centroid separation lower bound consists of two parts $E_1$ and $E_2$: $E_1$ reduced to the information-theoretically optimal threshold when $M=1$, corresponding to same covariance matrices; $E_2$ tends to vanish for small $M$ close to one and satisfying $M=1+o(1/\sqrt{n\log n})$ or remains as an extra term for large $M$. From our numerical results summarized in Figure~\ref{fig:sig_to_noise1}, we can observe that our defined centroid separation $\Delta$ indeed captures the accuracy of cluster label recovery using LA-SDP---the mis-clustering error curves display almost identical patterns under different settings of the GMM. In comparison, the performance of the (original) SDP~(\ref{eqn:kmeans_sdp}) and the $K$-means clustering method designed for the isotropic case become significantly worse as the condition number of the cluster covariance matrices increases. Here the mis-clustering error is defined as the ratio of mis-clustered data points to the total number of data points, which is minimized over all  permutations of the cluster labels. More details about implementation and model setups are provided in Appendix~\ref{app:exp}.

\subsection{Iterative LA-SDP under unknown covariance matrices: an alternating maximization algorithm}
\label{subsec:LA-SDP_feasible}

Since the oracle LA-SDP relies on the knowledge of covariance matrices $\Sigma_1, \dots, \Sigma_K$, we propose a simple and practical data-driven algorithm for approximating LA-SDP when these covariance matrices are unknown. The idea is to alternate between the SDP relaxation
given a current estimate of the component covariance matrices and updating covariance matrices according to the maximum (penalized) likelihood given the new membership estimate. The next lemma gives a closed-form formula for updating covariance matrices given a current estimate of the assignments $Z_1,\dots,Z_K$ based on their (unconstrained) MLEs on the observed data.

\begin{lem}[\bf Updating formula for covariance matrices under alternating maximization]\label{lem:assign_to_cov}
For any feasible matrices $Z_1, \dots, Z_K$ satisfying the constraints of~(\ref{eqn:kmeans_asdp}), the closed-form solution of the optimization problem \begin{equation}
\label{eqn:kmeans_asdp_inv}
\begin{gathered}
\hat{\Sigma}_1,\dots,\hat{\Sigma}_K = \arg\max_{\Sigma_1,\dots \Sigma_K \succeq 0} \sum_{k=1}^K\langle A_k, Z_k \rangle
\end{gathered}
\end{equation}
should be $\hat{\Sigma}_k:=$
\begin{equation}
    \frac{1}{\vone_{n}^T Z_k \vone_{n}}\sum_{i,j=1}^n \left[\frac{1}{2} (X_iX_i^T+X_jX_j^T)-X_iX_j^T\right] Z_{k,ij},
\end{equation}
$\forall k\in[K],$
where recall that $A^{(k)} := A^{(k)}(\Sigma_k)$ is the $\Sigma_k$-dependent similarity matrix defined in~(\ref{eqn:A_k}).
\end{lem}
Based on the lemma, we propose an \emph{iterative LA-SDP} (iLA-SDP) by alternating maximization of the profile log-likelihood~(\ref{eqn:profile_loglik}) for estimating the lifted cluster membership matrices $(Z_k)_{k=1}^K$ from LA-SDP~(\ref{eqn:kmeans_asdp}) and the component covariance matrices $(\Sigma_k)_{k=1}^K$, as summarized in Algorithm~\ref{alg:asdp}. The convergence of iLA-SDP for objective values can be directly implied from \ref{eqn:kmeans_asdp} and Lemma~\ref{lem:assign_to_cov}, where we summarize it as the theorem below.
\begin{thm}[\bf Monotonic maximization of iLA-SDP]\label{thm:lasdp}
The alternate updating rule of the assignments $Z_1^{(s)},\dots,Z_K^{(s)}$ and covariance matrices $\Sigma_1^{(s)},\dots,\Sigma_K^{(s)}$ in Algorithm~\ref{alg:asdp} will monotonically maximize the observed data log-likelihood over the iteration $s=0,1,2,\dots$, S.
\end{thm}

\begin{rem}
\label{rem:mono}
This theorem guarantees that the objective values are non-increasing over the iterations. However, obtaining an explicit theoretical convergence rate for iLA-SDP can be a challenging task.
\end{rem}
To understand the closed form solution of the covariance matrices in Lemma~\ref{lem:assign_to_cov}, we we first consider that in the special case where the lifted membership matrix $Z_k$ is of rank one, which holds for true lifted cluster membership matrices $(Z^\ast_k)_{k=1}^K$, the covariance matrices produced by iLA-SDP can be interpreted as within-cluster sample covariance matrices under soft clustering.


\begin{prop}[\bf Covariance estimation in iLA-SDP via soft clustering]\label{prop:soft_lasdp}
If $\rank(Z_k)=1,$ then there exists weights $(w_{k,1},\dots,w_{k,n})$ such that these $\hat{\Sigma}_k$ in Lemma~\ref{lem:assign_to_cov} can be written as
\begin{equation}
    \label{eqn:LA-SDP_updating_rule}
    \hat{\Sigma}_k:=\frac{1}{n_k}\sum_{i=1}^n w_{k,i} (X_i-\hat{\mu}_k)(X_i-\hat{\mu}_k)^\top,
\end{equation}
where $ \hat{\mu}_k:=\frac{1}{n_k} \sum_{i=1}^n w_{k,i}X_i \; \text{and} \; n_k = \sum_{i=1}^n w_{k,i}$. 
\end{prop}
It is further noted from the proof of Proposition~\ref{prop:soft_lasdp} that when $Z_k$ has rank one, the weights $w_{k,1}, \dots, w_{k,n}$ are proportional to the leading non-zero eigenvector of $Z_k$. Thus the alternating maximization step~(\ref{eqn:LA-SDP_updating_rule}) for updating the covariance matrices in iLA-SDP can be interpreted as a soft clustering technique that resembles the EM algorithm. Specifically, the E-step estimates the (hard) cluster label $Y_i \in \{0, 1\}^K$ associated with $X_i$ by the posterior probabilities $\tau_{ik} := p(Y_{ik} \mid X_i, \hat\theta^{(t)})$ where $\hat\theta^{(t)} = (\hat\pi_k^{(t)}, \hat\mu_k^{(t)}, \hat\Sigma_k^{(t)})_{k=1}^K$ denotes the estimated GMM parameters at the $t$-th iteration in the EM. Then the M-step updates the parameters via $\hat\pi_k^{(t+1)} = {m_k^{(t)} / n}$ with $m_k^{(t)} = \sum_{i=1}^n \tau_{ik}^{(t)}, \hat\mu_k^{(t+1)} = {1 \over m_k^{(t)}} \sum_{i=1}^n \tau_{ik}^{(t)} X_i,$
\begin{align*}
\hat\Sigma_k^{(t+1)}= {1 \over m_k^{(t)}} \sum_{i=1}^n \tau_{ik}^{(t)} (X_i - \hat\mu_k^{(t+1)}) (X_i - \hat\mu_k^{(t+1)})^\top. 
\stepcounter{equation}\tag{\theequation}\label{eqn:EM_Mstep}
\end{align*}
Note that~(\ref{eqn:LA-SDP_updating_rule}) and~(\ref{eqn:EM_Mstep}) represent different weighting schemes in the soft clustering rule for obtaining an estimate for the cluster labels. In iLA-SDP, the weight $w_{k,i}$ for $X_i$ belonging to component $k$ is determined by the SDP in~(\ref{eqn:kmeans_asdp}). Once the weights are calculated, remaining parameter updates in both iLA-SDP and EM boil down to simple averages with effective component sample sizes $n_k$ and $m_k$, respectively. In Section~\ref{subsec:connections} to follow, we provide deeper comparison between iLA-SDP and EM.

\begin{rem}
In Appendix~\ref{app:enhance_LASDP}, we further propose two variations of iLA-SDP that can handle high-dimensional and large-size data with better computational and statistical efficiency. For high-dimensional data, we apply Fisher's LDA with an initial estimate of the cluster labels to find an optimal feature subspace that increases the SNR for better clustering, and for large-size data we can use BM-like methods or combine the subsampling idea with iLA-SDP to reduce computational cost~\citep{pmlr-v151-zhuang22a}.
\end{rem}


\begin{algorithm}[h]
   \caption{The iterative likelihood adjusted SDP (iLA-SDP) algorithm}
   \label{alg:asdp}
\begin{algorithmic}[1]
   \STATE {\bfseries Input:} Data matrix $X\in\mathbb R^{p\times n}$ containing $n$ points. Initialization of assignments $G_1^{(0)},\dots,G_K^{(0)}$ or covariance matrices $\Sigma_1^{(0)},\dots,\Sigma_K^{(0)}$. The stopping criterion parameters $\epsilon$, $S$.
   \STATE (Assignments to covariance matrices) If we have the initialization of assignments, let  $\Sigma_k^{(0)}:={|G_k^{(0)}|}^{-1}\sum_{i\in G_k^{(0)} } (X_i-\bar{X}_k)(X_i-\bar{X}_k)^T$ to be the sample covariance of each cluster $k\in[K],$ where $\bar{X}_k:={|G_k^{(0)}|}^{-1}\sum_{i\in G_k^{(0)} } X_i. $
   \FOR{$s=1,\ldots,S$}
   \STATE   (Adjusted-SDP) Solve the Adjusted-SDP in~(\ref{eqn:kmeans_asdp}) using $X$ and $\Sigma_1^{(s-1)},\dots,\Sigma_K^{(s-1)} $ to get solution $Z_1^{(s)},\dots,Z_K^{(s)}$. 
      \STATE Compute the sum $\tilde{Z}^{(s)}:= \sum_{k=1}^K Z_k^{(s)}$ and the relative norm $r^{(s)}:=\|\tilde{Z}^{(s)}-\tilde{Z}^{(s-1)}\|_F/\|\tilde{Z}^{(s-1)}\|_F$ for $s\geq 2$. We will break the loop if $r^{(s)}<\epsilon$ .
       \STATE(Assignments to covariance matrices) Use formula in Lemma~\ref{lem:assign_to_cov} to get covariance matrices $\Sigma_1^{(s)},\dots,\Sigma_K^{(s)}$ from $Z_1^{(s)},\dots,Z_K^{(s)}$. 
   \ENDFOR
       \STATE Perform the spectral decomposition of $\tilde{Z}^{(S)}$ and take the top $K$ eigenvectors $(\hat{u}_1,\dots,\hat{u}_K)$. 
       \STATE Run $K$-means clustering on $(\hat{u}_1,\dots,\hat{u}_K)$ and extract the cluster labels $\hat{G}_1,\dots,\hat{G}_K$ as a partition estimate for $[n]$. 

   \STATE {\bfseries Output:} A partition estimate $\hat{G}_1,\dots,\hat{G}_K$ for $[n]$.
\end{algorithmic}
\end{algorithm}

\subsection{Connections between iLA-SDP and EM algorithms}
\label{subsec:connections}

It is interesting to observe that our proposed iLA-SDP algorithm is closely connected to the classic EM algorithm, which approximates the maximum likelihood estimation (MLE) of the observed data in statistical models with latent variables~\citep{EM1977}. The key idea of EM algorithm in the model-based clustering context is \emph{data augmentation} where the latent variables represent the cluster labels. More specifically, for each data point $X_i \in \bR^p$, we associate with an unobserved one-hot encoded cluster label $Y_i := \{Y_{i1}, \dots, Y_{iK}\} \in \{0, 1\}^K$. Then the EM algorithm aims to iteratively maximize the \emph{expected log-likelihood of the complete data} $(X_i, Y_i)_{i=1}^n$ given by $\theta^{(t+1)} :=$
\begin{equation}
    \label{eqn:EM_algo}
     \argmax_{\theta} \Big\{ Q(\theta \mid \theta^{(t)}) := \E_{\vY \sim q(\cdot \mid \vX, \theta^{(t)})}[ \ell_c(\theta \mid \vX, \vY) ] \Big\},
\end{equation}
where $\theta = ( (\pi_k, \mu_k, \Sigma_k)_{k=1}^K )$ contains parameters in the GMM, $(\pi_k)_{k=1}^K$ are the weight parameters such that $\pi_k \geq 0$ and $\sum_{k=1}^K \pi_k = 1$, and the complete log-likelihood function is
\begin{align*}
    &\ell_c(\theta \mid \vX, \vY) := p(\vX, \vY \mid \theta) = -{1\over2} \sum_{i=1}^n \sum_{k=1}^K Y_{ik} \cdot\\
    &\left[ \log(2\pi|\Sigma_k|) - (X_i - \mu_k)^\top \Sigma_k^{-1} (X_i - \mu_k) \right].
\end{align*}
Alternatively, the EM algorithm~(\ref{eqn:EM_algo}) can be interpreted as minorize-maximization (MM) that maximizes a best lower bound for the log-likelihood of the observed data
\begin{align*}
   & \ell(\theta \mid \vX) := \log p(\vX \mid \theta) = \log \sum_{\vY} p(\vX, \vY \mid \theta)\\
   &\geq \sum_{\vY} q(\vY \mid \vX) \log {p(\vX, \vY \mid \theta) \over q(\vY \mid \vX)} =: {\cal L}(q, \theta)
\end{align*}
for any posterior distribution $q(\vY \,|\, \vX)$. Under this perspective, the EM algorithm can be expressed as an \emph{alternating maximization} algorithm on ${\cal L}(q, \theta)$ between E-step $q^{(t+1)} = \argmax_q {\cal L}(q, \theta^{(t)})$ and M-step $\theta^{(t+1)} = \argmax_\theta {\cal L}(q^{(t+1)}, \theta)$. Thus, give any $q(\vY \,|\, \vX)$, the M-step maximizes the expected complete log-likelihood as a surrogate function that minorizes $\ell(\theta\,|\,\vX)$ because ${\cal L}(q, \theta) = \E_{\vY \sim q(\cdot \,|\, \vX)} [\ell_c(\theta \,|\, \vX, \vY)] - H(q(\vY \,|\, \vX))$ where $H(q)$ denotes the relative entropy of distribution $q$, while given the current parameter estimate $\theta^{(t)}$, the E-step is maximized at $q^{(t+1)}(\vY \,|\, \vX) = p(\vY \,|\, \vX, \theta^{(t)})$ because
\begin{align*}
   & \ell(\theta^{(t)} \mid \vX) \geq {\cal L}(p(\vY \mid \vX, \theta^{(t)}), \theta^{(t)}) \\
   &= \sum_{\vY} p(\vY \mid \vX, \theta^{(t)}) \log p(\vX \mid \theta^{(t)}) = \ell(\theta^{(t)} \mid \vX),\\[-1.5ex]
\end{align*}
where the first inequality is actually an equality at $p(\vY \,|\, \vX, \theta^{(t)})$. Even though the EM and iLA-SDP are both alternating maximization algorithms aiming to solve the MLE for the observed data log-likelihood and both can be viewed as soft clustering methods (cf. Proposition~\ref{prop:soft_lasdp}), there are several important differences we would like to highlight.  
    
\begin{figure}[t] 
   \centering
   \vspace{0.1cm}
     \subfigure{\includegraphics[trim={1.1cm 7cm 1.1cm 6cm},clip,scale=0.4]{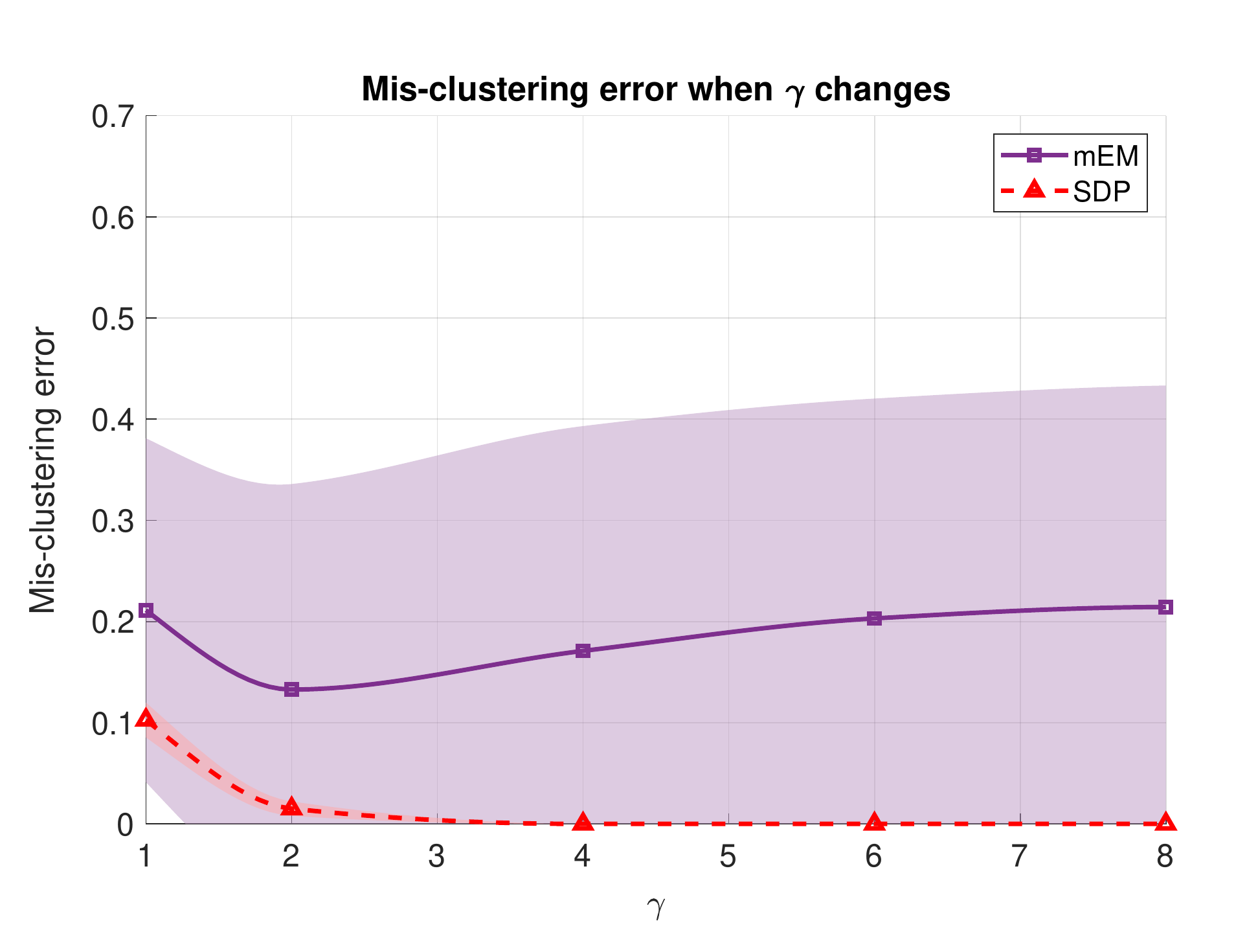}} 
     \subfigure{\includegraphics[trim={1.1cm 5.5cm 1.1cm 7cm},clip,scale=0.4]{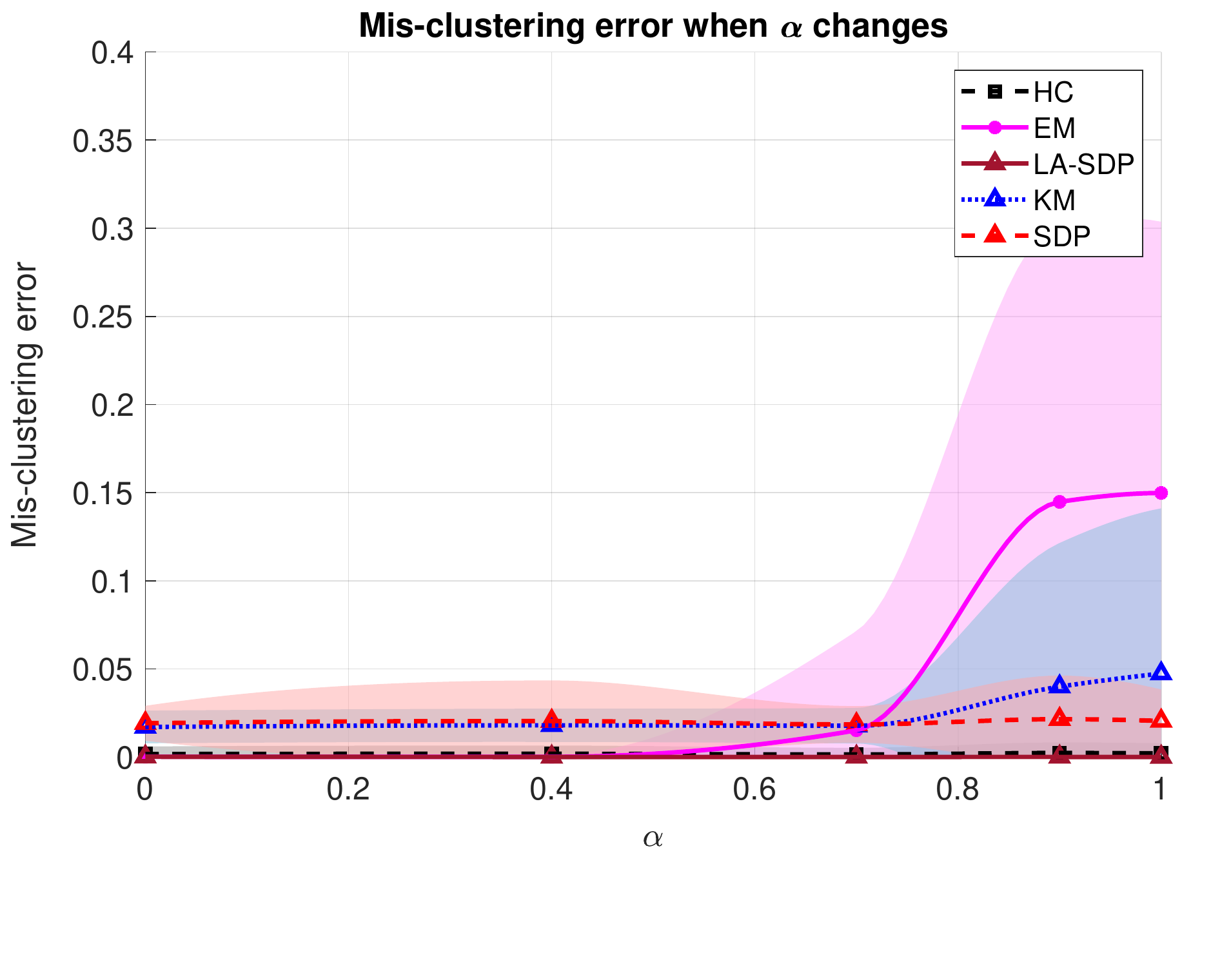}}\\[-7ex]
   \caption{Mis-clustering error (with shaded error bars) vs $\gamma$ (captures the signal strength of GMM) and $\alpha$ (perturbation percentage of initialization). mEM (SDP) refers to the reduced version of EM (LA-SDP) where we consider covariance matrices as fixed and equal to identity. The first plot compares the performance of mEM and SDP when separation is large with random initialization; the second plot compares all methods when we enlarge the perturbation percentage $\alpha$ applied to the random initialization from hierarchical clustering (HC). }
   \label{fig:error_rates}
\end{figure}

    First, cluster labels are (random) latent variables and they are estimated via posterior probabilities in the EM algorithm, while the labels are treated as unknown parameters in iLA-SDP that are estimated via direct maximization of  the observed data likelihood.
    
    Second, the EM algorithm is a special case of the minorization-maximization (MM) algorithm~\citep{HunterLange2000_MM-algo} by iteratively performing the coordinate ascent on the expected complete data log-likelihood as a minorizing surrogate function, while our iLA-SDP is \emph{exact} in the sense that it directly optimizes the observed data log-likelihood via a convex relaxation formulation. Thus iLA-SDP is a more direct approach than EM for tackling the non-convex observed log-likelihood objective and it is principled to perfectly recover the true clustering structure if the clusters are well-separated under an SNR lower bound in Theorem~\ref{thm:main_thm}. As in the EM algorithm, iLA-SDP monotonically maximizes the observed data log-likelihood over iterations, where we can get both theoretically from Theorem~\ref{thm:lasdp} and empirically; cf. Figure~\ref{fig:loglik} in Appendix.

    Third, the EM algorithm in each iteration must estimate the cluster center parameters $(\mu_k)_{k=1}^K$, while our iLA-SDP profiles out the effect of centroid estimation and leverages only pairwise Mahalanobis distances between data to accommodate the heterogeneity of cluster shapes. Partly because the error in estimating the centroids propagates to other parameters, EM is more sensitive to initialization with inaccurate labels and the centroid configurations even in the standard GMM~\citep{JinZhangSilvaramanWainwrightJordan2016_EM}, and iLA-SDP behaves better than EM, an observation we empirically verify in our simulation experiments; cf. Figure~\ref{fig:error_rates} for comparison between iLA-SPD and EM algorithms.

  {\bf Failure of EM vs SDP.} The failure of EM for random initialization~\citep{JinZhangSilvaramanWainwrightJordan2016_EM} in the special case that covariance matrices equal to identity matrix and it assumes equal weights. Both covariance matrices and weights are known. In this case, EM algorithm would be reduced to the version that the weights and the mean update interactively. Meanwhile, LA-SDP would be reduced to SDP. The random initialization indicates that we pick any data point as initialization of the centers uniformly. Following the same setting from the construction of the pitfall, we choose one dimension GMM with three clusters such that the distance between two of the centers is much smaller than others. More concisely, we let $n=300,\; K=3,\;p=1,\;\mu_1=\gamma,\;\mu_2=-\gamma,\;\mu_3=10\cdot\gamma.$ The results can be observed from the first plot in Figure~\ref{fig:error_rates} with $300$ replicates, where we denote the reduced version of EM as mEM. From the first plot we can observe that LA-SDP with isotropic known covariance matrices, which reduces to the $K$-means SDP in~(\ref{eqn:kmeans_sdp}), performs stable and achieves exact recovery when the separation is large. However, EM  fails with random initialization in this adversarial centroids configuration. 

    {\bf Perturbation of initialization assignments.} To see how the performance of EM and LA-SDP will change when perturbing the initialization, we set HC as initialization and proportion $\alpha$ ($\alpha\in[0,1]$) of the initialization labels will be perturbed. The diagonal of the covariance matrices are placed at a simplex of $\bR^p$ that are not identical to the corresponding centers. i.e. $\mu_k=\lambda\cdot e_k$, $\Sigma_k=L\cdot \diag(e_{k+1}),\; \forall l\in[K],$ where $e_{K+1}=e_1.$ This guarantees the symmetry of the construction. We set $L=10,\; p=4,\; K=4$ and the distance between centers $d=8.$ Each time we draw the $n=200$ data from the GMM and run HC as initialization. Then we randomly assign $\alpha$ proportion of the labels from HC to any cluster uniformly. The results of the simulation for the second plot in Figure~\ref{fig:error_rates} are obtained through $300$ total replicates,
    where we can see that LA-SDP is fairly stable with perturbation of initialization if the separation is large while EM can go worse as the perturbation percentage of initialization $\alpha$ approaches 1, i.e., all the labels are selected randomly. In other words, EM is more sensitive to initialization and LA-SDP is more stable if the signal is strong. More discussions about iLA-SDP can be found in Appendix~\ref{app:dis}.

\section{Real-data applications}\label{sec:real_data}
\begin{figure}[!b] 
   \centering
   \vspace{0.1cm}
      \subfigure{\includegraphics[trim={0.5cm 7cm 1.45cm 7cm},clip,scale=0.42]{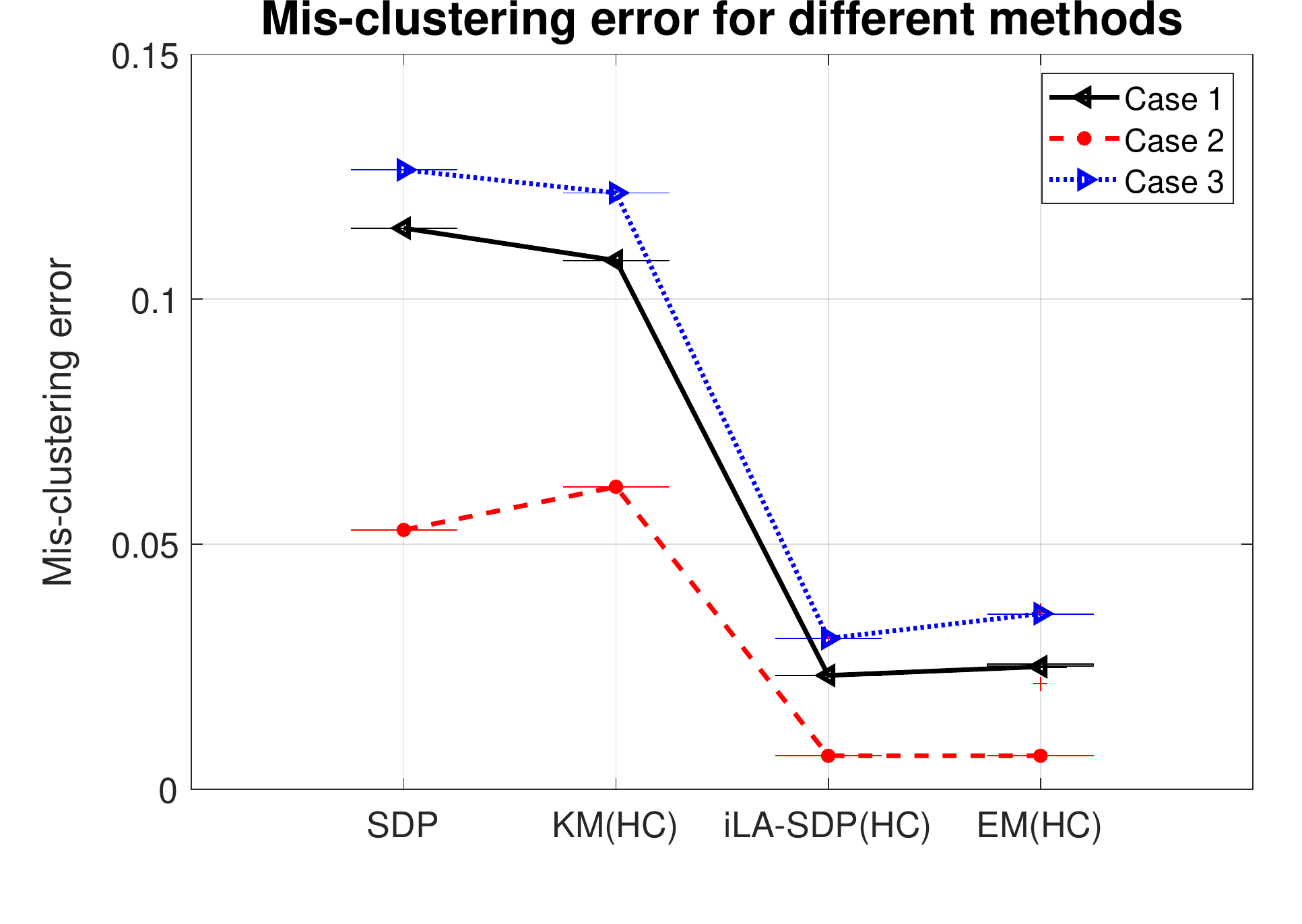}} 
      \subfigure{\includegraphics[trim={0.5cm 6.5cm 1.0cm 7cm},clip,scale=0.42]{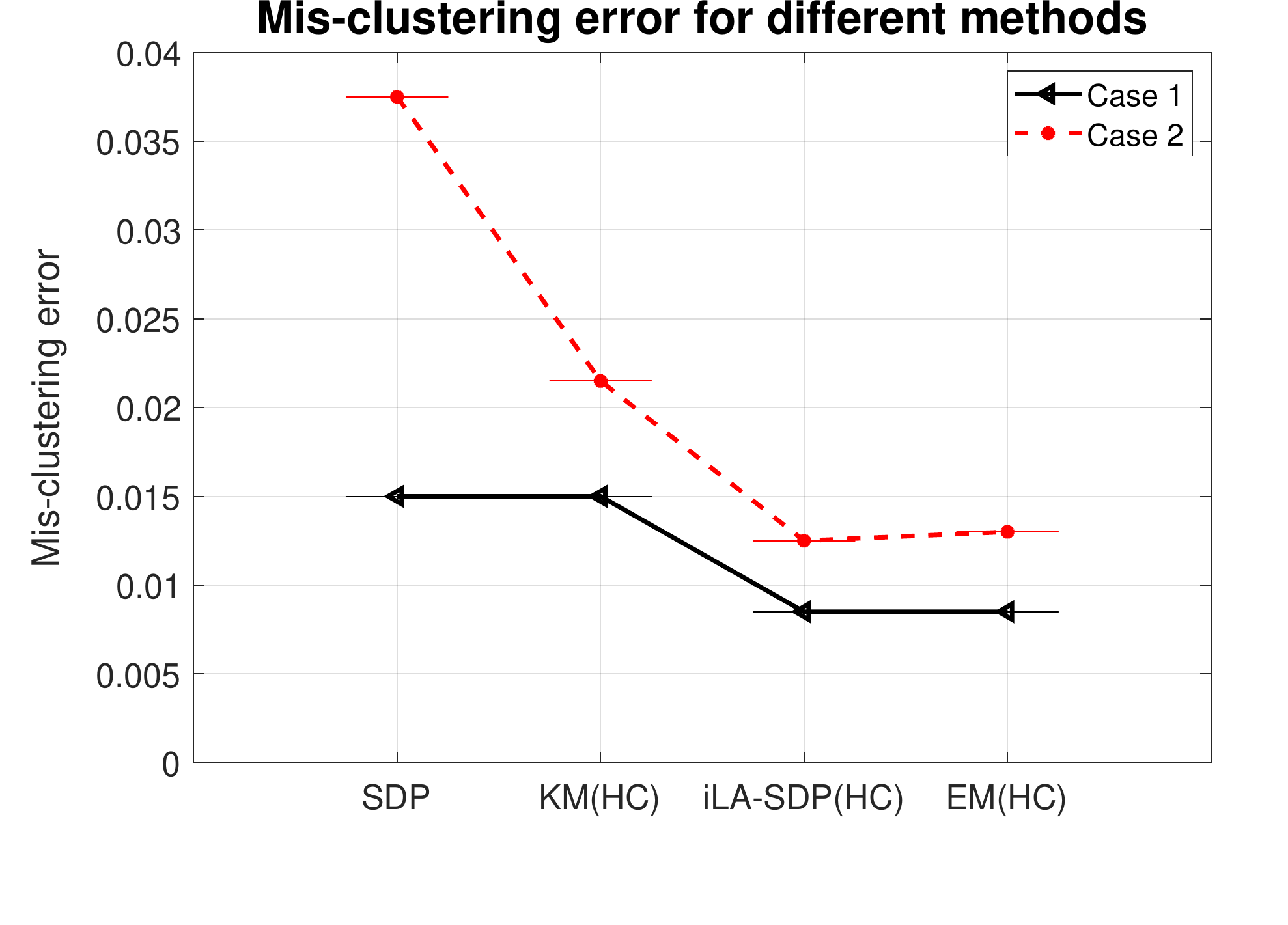}}
      \subfigure{\includegraphics[trim={0.5cm 5.5cm 1.0cm 7cm},clip,scale=0.42]{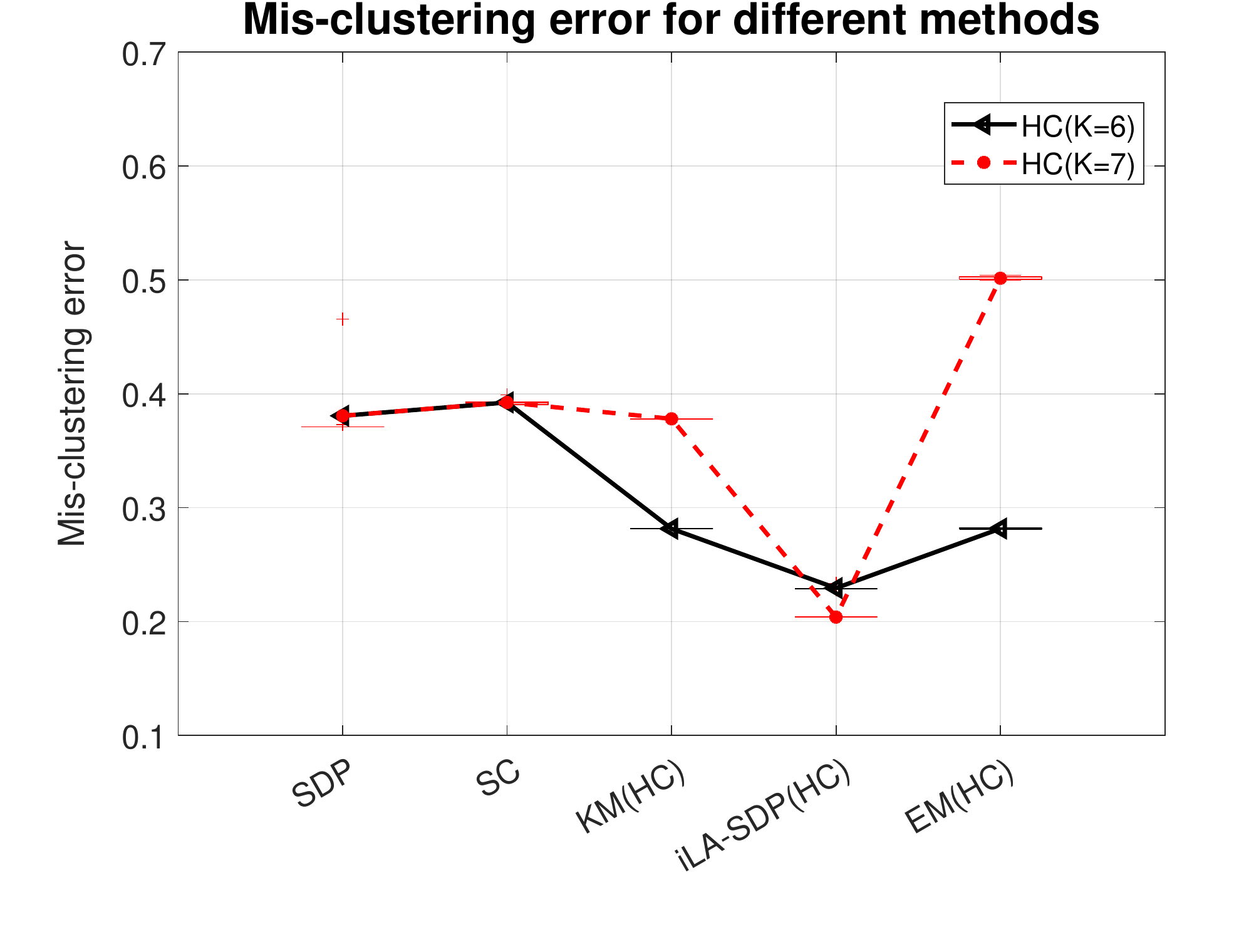}}\\[-7ex]
      \vspace{0.2cm}
\caption{Boxplots of mis-clustering error (with means) for different methods. The top (middle) plot summarizes the results for the MNIST dataset (CIFAR-10 dataset). The bottom plot summarizes the results for the Landsat satellite dataset.
}
   \label{fig:barplot}
\end{figure}

In this section, we test the performance of iLA-SDP against several widely used clustering methods on three real
datasets. The first one is handwriting digits dataset MNIST, the second one is CIFAR-10 image dataset and the last one is Landsat dataset from the UCI machine learning repository. The detailed settings can be found in Appendix~\ref{appendix:app}. As we know, the time complexity of solving SDP (iLA-SDP) is of order $O(n^{3.5})$. To resolve that issue, in practice, we used Burer Monteiro approach (BM method) to bring down the time and space complexity to nearly $O(n)$ with enough accuracy; cf. Figure~\ref{fig:logtime} in Appendix~\ref{app:exp}. Specifically, we adapted the BM formulation for the SDP~\citep{sahin2022inexact} to suit our LA-SDP (\ref{eqn:kmeans_asdp}) setting, i.e.,
\begin{align*}
&\big(\hat{U}_1,\dots,\hat{U}_K\big) = \argmax_{U_1,\dots U_K \in \mathbb{R}^{n \times s}} \sum_{k=1}^K\langle A_k, U_kU_k^T \rangle, \\
&\mbox{subject to }   \|U\|^2 = K,\;UU^T \vone_{n} = \vone_{n}, \;U \geq 0,\\
\stepcounter{equation}\tag{\theequation}
& U=[U_1 |\dots|U_K]\in \bR^{n\times r},\; r= sK\geq K.
\label{eqn:kmeans_asdp_BM}
\end{align*}
 The details of the algorithm are beyond the scope of this topic and will be elaborated in our following works.

{\bf Handwriting digits dataset.} First let us look at the performances of our methods for a handwriting digits dataset MNIST. We used PCA to extract features from the training set and reduce the dimension for the test set to $q=40$. The given test set contains $10000$ samples with $K=10$ clusters. We choose three clusters with $n$ roughly equal $3000$ and perform clustering algorithms for total $10$ replicates. Case 1, case 2 and case 3 correspond to choosing digits "0", "2" and "3"; digits "3", "4" and "6" and digits "3", "4" and "8" respectively. HC is used as initialization for EM, KM and iLA-SDP. The comparison of those four methods can be found on the top in Figure~\ref{fig:barplot}, where we can observe that iLA-SDP and EM in this case have comparable behaviors. They both achieve better performance than KM and SDP. The results of SC (spectral clustering) have not been included in this plot as SC failed with mis-clustering error larger that $0.5$ for this situation. Over the experiments we found the comparable behaviors for EM and iLA-SDP, this is reasonable since their goals are both minimizing the log-likelihood function. When the signal is strong, both methods outperform KM and SDP that only consider fixed covariance matrices as we suspected. 

{\bf CIFAR-10 dataset.} Next we perform all methods for CIFAR-10 dataset. This dataset consist of 60000 $32\times 32 \times 3$ colored images with 10 clusters.  We used Inception v3 model and default settings to finally extract 2048 features, after which we used PCA to further reduce the dimension for the test set to $q=20$. The given test set contains $10000$ samples with $K=10$ clusters. We simply choose two clusters of them with sample size $n = 2000$. The results are shown in the second plot of Figure~\ref{fig:barplot} with 10 replicates. Case 1 (case 2) corresponds to choosing "dog" and "ship" ("automobile" and "horse") respectively. The results of spectral clustering failed in this situation. Similar to previous experiment, we can see that iLA-SDP and EM both achieve relatively good performance comparing to KM and SDP under those circumstances.

{\bf Landsat satellite dataset.} This database was generated from landsat Multi-Spectral Scanner image data. The test set includes $2000$ satellite images, $6$ different clusters with $36$ attributes ($36 = 4$ spectral bands $\times$ $9$ pixels in neighbourhood). Every attribute is an integer from $0$ to $255$ indicating the color for certain pixel. We performed 5 methods on the transformed dataset with total $10$ replicates. For each attribute, we scale its range to $[0,1]$ and then take the function $f(x)=\log(1/x-1)$ entry-wise to transform the range to $\bR_+$. Then, we run Algorithm~\ref{alg:asdp_hd2} on the transformed dataset $\tilde{X}$ to get the results for LA-SDP with $ q=K=6$. We ensure the randomness only comes from initialization (and the rounding procedure) to see how initialization affects the performance of different methods. From the results we can see that our method LA-SDP performs the best for all four methods. The initialization for KM, iLA-SDP and EM is hierarchical clustering. Especially, if we attack the initialization by setting $K=7$ (originally $K=6$) in HC, iLA-SDP performed stable while EM failed. Because there are both biases of the estimations of group means and covariance matrices for EM while iLA-SDP only uses the group covariance matrices information as its initialization. So when the information of initialization, especially for the group means, got jeopardized, KM and EM will be affected easily while iLA-SDP should be stable relatively. We can observe that the mis-clustering error for SDP has larger variance (outliers) in this case, this is because the final assignments are obtained from rounding process. The results from SDP are far from exact recovery, so the rounding process may induce more variants.

\section{Discussion}\label{sec:discuss}
One limitation of iLA-SDP is its reliance on the assumption that all covariance matrices of clusters are non-singular --- if any of the true covariance matrices happens to be singular, the method becomes inapplicable. In such cases, additional adjustments such as projecting the data into suitable lower-dimensional subspaces would be necessary.  There are other clustering methods that are less sensitive to the initialization such as power $K$-means~\cite{pmlr-v97-xu19a}. However, objective function of this method does not take the second-order information and thus we expect it would be inferior comparing to EM or iLA-SDP if heterogeneity is present in data.  In summary, iLA-SDP combines the strengths of both SDP (robustness to initialization and against outliers / adversarial attack) and EM (flexibility to estimate cluster-specific covariance matrices). By leveraging the Burer-Monteiro approach~\citep{BurerMonteiro2003}, the per-iteration computational complexity of iLA-SDP is similar to that of EM, both scaling at $O(n)$.

\section*{Acknowledgements}

Xiaohui Chen was partially supported by NSF CAREER grant DMS-1752614. Yun Yang was partially
supported by NSF grant DMS-2210717.

\newpage
\bibliography{icml2023}
\bibliographystyle{icml2023}

\newpage
\appendix
\onecolumn

\section{Appendix}

\label{app:review}
\begin{figure}[h!] 
   \centering
   \includegraphics[trim={1.45cm 8cm 1.45cm 8cm},clip,scale=0.6]{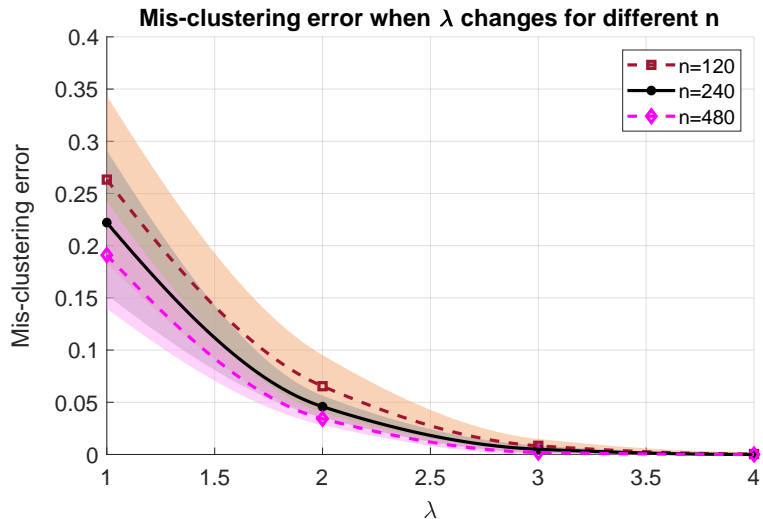}
 
   \caption{Mis-clustering error (with shaded error bars for the left plot) vs $\lambda$ for LA-SDP for different $n$. }
   \label{fig:oracle}
\end{figure}

{\bf Sample complexity bound.} To verify the sample complexity bound for LA-SDP in Theorem~\ref{thm:main_thm} ($O(\log(n))$) is tight, we will change $n$ and adjust the squared distance between clusters by multiplying $\log(n)$. More precisely, we let $d= \lambda\sqrt{\log(n)},\; \lambda>0.$ The diagonal of the covariance matrices are placed at a simplex of $\bR^p$ that are not identical to the corresponding centers. i.e. $\mu_k=\lambda\cdot e_k$, $\Sigma_k=L\cdot \diag(e_{k+1}),\; \forall l\in[K],$ where $e_{K+1}=e_1.$ This guarantees the symmetry of the construction. We set $L=10,\; p=4,\; K=4$. Each time we draw the $n=120/240/480$ data from the GMM. The results of the simulation for the second plot in Figure~\ref{fig:oracle} are obtained through $20$ total replicates, where we can observe the same pattern across different settings for $n.$ This shows that the order $\log(n)$ for separation bound in Theorem~\ref{thm:main_thm} should be tight.

{\bf Related works.} Under the Gaussian mixture model (GMM) with isotropic noise, $K$-means clustering is equivalent to the maximum likelihood estimator (MLE) for cluster labels, which is known to be worst-case NP-hard~\citep{aloise2009np}. Fast approximation algorithms to solve the $K$-means such as Lloyd's algorithm~\citep{Lloyd1982_TIT,LuZhou2016} and spectral clustering~\citep{Meila01learningsegmentation,NgJordanWeiss2001_NIPS,Vempala04aspectral,Achlioptas2005McSherry,vanLuxburg2007_spectralclustering,vonLuxburgBelkinBousquet2008_AoS} provably yield consistent recovery when different groups are well separated. Recently, semi-definite programming (SDP) relaxations~\citep{PengWei2007_SIAMJOPTIM,MixonVillarWard2016,LiLiLingStohmerWei2017,FeiChen2018,CHEN2021303,Royer2017_NIPS,GiraudVerzelen2018,BuneaGiraudRoyerVerzelen2016} have emerged as an important approach for clustering due to its superior empirical performance~\citep{PengWei2007_SIAMJOPTIM}, robustness against outliers and adversarial attack~\citep{FeiChen2018}, and attainment of the information-theoretic limit~\citep{chen2021cutoff}. Despite having polynomial time complexity, the SDP relaxed $K$-means has notoriously poor scalability to large (or even moderate) datasets for instance by interior point methods~\citep{Alizadeh1995,Jiang2020_FOCS}, as the typical runtime complexity of an interior point algorithm for solving the SDP is at least $O(n^{3.5})$, where $n$ is the sample size.

\subsection{Enhanced iLA-SDPs for high-dimensional and large-size data}
\label{app:enhance_LASDP}
In this section, we propose two variations of iLA-SDP that can handle high-dimensional and large-size data with better computational and statistical efficiency.

{\bf High dimensional data.} If the number attributes of the data are large, it would be hard to approximate the true covariance matrices since there are $O(p^2)$ many unknown parameters. Thus, we propose two dimension reduction procedures that based on hierarchical clustering, Fisher's LDA and F-test. The detailed algorithm have been shown in Algorithm~\ref{alg:asdp_hd1} and Algorithm~\ref{alg:asdp_hd2}. To reduce the dimension, we proposed two procedure. 
\begin{enumerate}[leftmargin=*]
    \item If the number of clusters $K$ is small and the difference between centers are sparse, we shall use HC as a benchmark method for feature selection and assume the group means according to HC as ground true. Specifically, for $i$-th attribute, we calculate the F-statistics and its p-value based on the $H_0$ that all group means w.r.t. $i$-th attribute are the same. At last, each attribute would likely to be selected if the p-value $\mathcal{P}_i$ for $i$-th attribute is significantly small among p-values for all attributes. 
    
    \item First we use the hierarchical clustering to get the clustering results for all possible input cluster number $\tilde{K}\in[p].$ If we assume all the clusters have identical covariance matrices, then we may use the assignments from HC to estimate the within-cluster covariance $\tilde{W}$ (with group means $\tilde{\mu}_l$) and get the signal-to-noise ratio $\Delta(\tilde{K}):=\min_{k\ne l} \|\tilde{W}^{-1/2} (\tilde{\mu}_k-\tilde{\mu}_l) \|$. Here, HC serves as a benchmark method for data initial processing. We will then choose the largest $\tilde{K}$ within target range such that the signal-to-noise ratio $\Delta(\tilde{K})$ is maximized. Then it will lead to the new dataset with dimension $q=\tilde{K}-1$ after running Fisher's LDA on the assignments from HC with clusters number equals $\tilde{K}$. Finally we perform Algorithm~\ref{alg:asdp} on the new dataset and extract the cluster labels.

\end{enumerate}

{\bf Large-size data.} As we know that the time complexity for solving SDP is as high as $O(n^{3.5}).$ We used the BM methods in practice. The details will be fully argued in following workds. One might also use subsampling methods to bring down the time cost while maintain the superior behavior for LA-SDP~\citep{pmlr-v151-zhuang22a}. The proposed algorithm is shown in Algorithm~\ref{alg:asdp_n}.

\begin{algorithm}[h]
   \caption{Likelihood adjusted SDP based iterative algorithm  with unknown covariance matrices  $\Sigma_1,\dots,\Sigma_K$ for large $p$.}
   \label{alg:asdp_hd1}
\begin{algorithmic}[1]
   \STATE {\bfseries Input:}Data matrix $X\in\mathbb R^{p\times n}$ containing $n$ points. Cluster numbers $K$. The stopping criterion parameters $p_0$, $\epsilon$ and $S$. $\alpha\in [0,1],\; C>0.$
   \STATE Run hierarchical clustering with data $X$, clusters number $K$ and extract the cluster labels $G_1^{(0)},\dots,G_{K}^{(0)}$ as prior assignments for $[n]$. Suppose the assignments have true centers $\mu_{k}^{(0)},\; k\in[K]$.
   \FOR{$i=1,\dots,p$}
   \STATE   Calculate the p-value $\mathcal{P}_i$ of the F-test $\mathcal{F}_i$ under $H_0$: $\mu_{1,i}^{(0)}=\dots=\mu_{K,i}^{(0)},$ where  $\mu_{k,i}^{(0)}$ corresponds to the $i$-th component of $\mu_{i}^{(0)}$.
   \ENDFOR
      \STATE Keep $p_0$ attributes with $p_0$ smallest p-values $\mathcal{P}_i$.
       \IF{there is no clear cutoff between $\mathcal{P}_i$'s, i.e. $\max_{i\in[p]}\mathcal{P}_i/\min_{i\in[p]}\mathcal{P}_i<C,$}
       \STATE we further keep other $p-p_0$ attributes with probability $\alpha>0.$
       \ENDIF
       \STATE Get dimension reduced data $\tilde{X}$.
       \STATE Run Algorithm~\ref{alg:asdp} on $\tilde{X}$ with initialization obtained from $K$ clusters of HC and stopping criterion parameters $\epsilon$ and $S$. Then extract the cluster labels $\hat{G}_1,\dots,\hat{G}_K$ as a partition estimate for $[n]$. 
   \STATE {\bfseries Output:}A partition estimate $\hat{G}_1,\dots,\hat{G}_K$ for $[n]$.
\end{algorithmic}
\end{algorithm}
 
\begin{algorithm}[h]
   \caption{Likelihood adjusted SDP based iterative algorithm  with unknown covariance matrices  $\Sigma_1,\dots,\Sigma_K$ for large $p$.}
   \label{alg:asdp_hd2}
\begin{algorithmic}[1]
   \STATE {\bfseries Input:}Data matrix $X\in\mathbb R^{p\times n}$ containing $n$ points. Cluster numbers $K$, the reduction dimension $\tilde{K}\in [K,p]$. The stopping criterion parameters  $\epsilon$ and $S$.
   \STATE Select a bench mark clustering method (HC) as a way to provide a prior assignments.
\STATE Choose the reduction dimension $\tilde{K}\in [K,p]$. Run hierarchical clustering with data $X$, clusters number $\tilde{K}$ and extract the cluster labels $G_1^{(\tilde{K})},\dots,G_{\tilde{K}}^{(\tilde{K})}$ as prior assignments for $[n]$.
   \STATE   Perform the Fisher's LDA with data $X$, assignments $G_1^{(\tilde{K})},\dots,G_{\tilde{K}}^{(\tilde{K})}$ and get the transformed data $\tilde{X}\in \mathbb R^{q\times n}$ with $q=\tilde{K}-1 $.
      \STATE Run Algorithm~\ref{alg:asdp} on $\tilde{X}$ with initialization obtained from $K$ clusters of HC and stopping criterion parameters $\epsilon$ and $S$. Then extract the cluster labels $\hat{G}_1,\dots,\hat{G}_K$ as a partition estimate for $[n]$.
   \STATE {\bfseries Output:}A partition estimate $\hat{G}_1,\dots,\hat{G}_K$ for $[n]$.
\end{algorithmic}
\end{algorithm}

\begin{algorithm}[h]
   \caption{Sketch and lift: Likelihood adjusted SDP based iterative algorithm  with unknown covariance matrices  $\Sigma_1,\dots,\Sigma_K$ for large $n$.}
   \label{alg:asdp_n}
\begin{algorithmic}[1]
   \STATE {\bfseries Input:}Data matrix $X\in\mathbb R^{p\times n}$ containing $n$ points. Cluster numbers $K$. The stopping criterion parameters $P$, $\epsilon$ and $S$. Sampling weights $(w_1,\dots,w_n)$  with $w_1 = \dots = w_n = \gamma \in (0,1)$ being the subsampling factor.
   \STATE (Sketch) Independent sample an index subset $T \subset [n]$ via $\text{Ber}(w_i)$ and store the subsampled data matrix $V = (X_i)_{i \in T}$.
 \STATE Run subroutine Algorithm~\ref{alg:asdp} with input $V$ to get a partition estimate $\hat{R}_1,\dots,\hat{R}_K$ for $T$.
\STATE   Compute the centroids $\bar{X}_k = |\hat{R}_k|^{-1} \sum_{j \in \hat{R}_k} X_j$ and within-group sample covariance matrices $\hat{\Sigma}_k = |\hat{R}_k|^{-1} \sum_{j \in \hat{R}_k} (X_j-\bar{X}_k)(X_j-\bar{X}_k)^T$ for $k \in [K]$. 
   \STATE   (Lift) For each $i \in [n] \setminus T$, assign $i \in \hat{G}_k$ if
      \STATE $\log |\hat{\Sigma}_k|+\|\hat{\Sigma}_k^{-1/2}(X_i - \bar{X}_k)\|^2 <\log |\hat{\Sigma}_l|+ \|\hat{\Sigma}_l^{-1/2}(X_i - \bar{X}_l)\|^2, \quad \forall l \neq k,\, l \in [K]$. And randomly assign $i$ to any $K$ clusters if such $k$ doesn't exist.
   \STATE {\bfseries Output:}A partition estimate $\hat{G}_1,\dots,\hat{G}_K$ for $[n]$.
\end{algorithmic}
\end{algorithm}

\subsection{Experiment results}
\label{app:exp}
In this section, we provide more details of the settings and post the results for simulation experiments. For all the dimension reduction procedures used in the simulation experiments, we perform step 1-7 in Algorithm~\ref{alg:asdp_hd1} followed by Algorithm~\ref{alg:asdp_hd2} with input parameters $\alpha=0.7,\; C=10^{10},\;p_0=2K,\; p_1=15\;\epsilon=10^{-2},\; S=50.$ The initialization we use is hierarchical clustering from \emph{mclust} package in R. Here we test our algorithm on Gaussian mixture models and real datasets. We compared our algorithm LA-SDP (HC as initialization) with HC, EM algorithm (HC as initialization), $K$-means (HC as initialization) and original SDP. 

{\bf Improvements of LA-SDP over SDP.} Recall in Theorem~\ref{thm:main_thm}, we define the signal-to-noise ratio as $\Delta^2:= \min_{k\ne l}\|\Sigma_k^{-1/2}(\mu_k-\mu_l) \|^2 $. To verify the validity of the definition and compare LA-SDP and SDP, we change the conditional number for covariance matrices $\Sigma_1,\dots,\Sigma_K.$ Here we choose $n=200,\; p=4,\; K=4.$  Recall $M:=\max_{k \ne l}\|\Sigma_l^{1/2}\Sigma_k^{-1}\Sigma_l^{1/2} \|_{\rm op} $, we choose all the covariance matrices to be the same such that $M$ is fixed. The covariance matrices are set to be identity matrix except that the first entry at the diagonal are set to be $L+1$, which refers to the condition number of matrices. We consider two cases where $L=10,\;100.$ Now denote $e_k\in\bR^p$ as the vector with $k$-th entry as $1,$ and $0$ otherwise. The centers of clusters $\mu_1,\dots,\mu_K$ are placed on vertices of a regular simplex, i.e., $\mu_k=\lambda \sqrt{1+(1+L)^{-1}} e_k, \; k\in [K].$ This ensures that for any $L$, $ \Delta=\lambda,\; \forall \lambda.$ From Figures~\ref{fig:sig_to_noise1} we can observe that the signal-to-noise ratio we defined is reasonable. On the other hand, the performance of SDP becomes worse as condition number of the group covariance matrices grows since the assumption of isotropy group covariance matrices for SDP is violated and same reason for $K$-means.

\begin{figure}[h!] 
   \centering
   \includegraphics[trim={1.45cm 7cm 1.45cm 7cm},clip,scale=0.53]{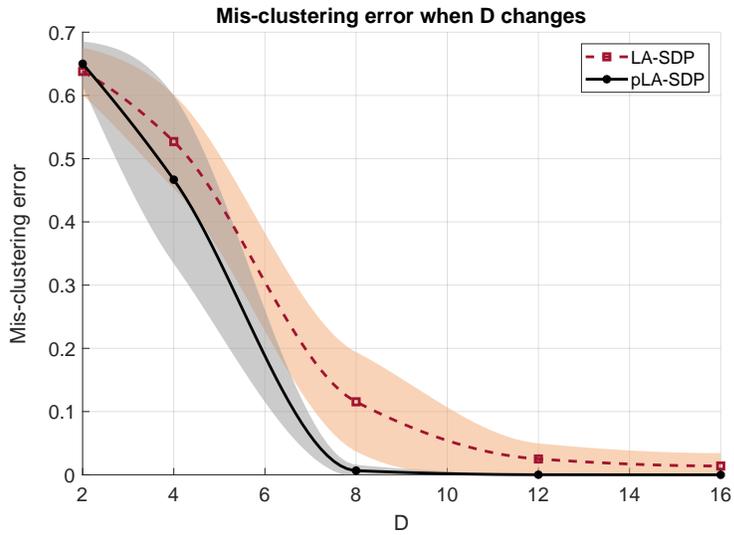}
 
   \caption{Mis-clustering error (with shaded error bars for the left plot) vs center distance $D$ for LA-SDP before and after dimension reduction. pLA-SDP denotes the LA-SDP after dimension reduction. }
   \label{fig:dim_reduce}
\end{figure}

{\bf Impact of dimension reduction.} 
Here we want to see the performance of LA-SDP after dimension reduction. The covariance matrices of GMM are drawn independently following $\Sigma_k:=U_k\Lambda_k U_k^T,\; \forall k\in[K].$ Here $U_k$ is a random orthogonal matrix, $\Lambda_k$ is a diagonal matrix with entries drawn from $\mathcal{Z}=1+\beta Z\cdot\vone(Z>0),$ where $Z$ is standard Gaussian distribution, $\beta>0$ controls the condition number of $\Sigma_k$. Here we choose $n=200,\; p=20,\; K=4,\;\beta=5.$ The covariance matrices are fixed once chosen and we perform Algorithm~\ref{alg:asdp} on the dataset directly to get the results of LA-SDP for each replicates. For dimension reduction, we follow the procedure of dimension reduction introduced in Algorithm~\ref{alg:asdp_hd1} and Algorithm~\ref{alg:asdp_hd2} in Appendix~\ref{app:enhance_LASDP} and get the transformed dataset $\tilde{X}$ with lower dimension. Then the results of pLA-SDP is obtained from running Algorithm~\ref{alg:asdp} with HC as initialization on $\tilde{X}$. The results in Figure~\ref{fig:dim_reduce} shows that after reduction of dimension in our procedure, the performance of LA-SDP becomes significantly better when the separation is large. This is because in our setting, the difference between centers $d_{(k,l)}:=\mu_k-\mu_l,$ is sparse for all distinct pairs. And after performing the F-test on the covariates, the noisy terms get eliminated which results in better performance.

{\bf Empirical evidence for ADMM of LA-SDP.} Here we provide examples based on previous experiment settings where we set the distance between centers $d=1/3/5/10.$  and try to see how the log-likelihood function of given data changes as the iteration proceeds. From Figure~\ref{fig:loglik} in Appendix we can see that our algorithm guarantees that the log-likelihood function of given data increases over iteration empirically. What is more, by our construction we can show that the log-likelihood function will increase after each step of ADMM for LA-SDP theoretically.

\begin{figure}[t] 
   \vspace{0.1cm}
   \centering
   \includegraphics[trim={0.01cm 7cm 0.01cm 7cm},clip,scale=0.55]{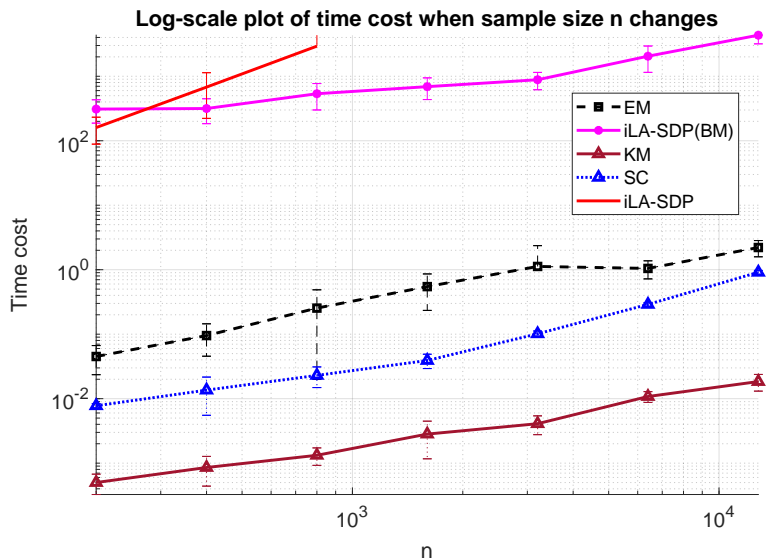}
   \vspace{0.1cm}
   \caption{Log-scale plot of time cost (with error bars) vs sample size $n$. iLA-SDP(BM) corresponds to our algorithm iLA-SDP based on BM; iLA-SDP uses SDP solver directly; KM corresponds to $K$-means and SC corresponds to spectral clustering.}
   \label{fig:logtime}
\end{figure}

{\bf Computational complexity of iLA-SDP (based on Burer Monteiro approach).} Our BM formulation of LA-SDP has been expressed as~(\ref{eqn:kmeans_asdp_BM}).
We have plotted the log-scale of time cost versus sample complexity for various methods with 10 replicates, based on the first experiment's setting. Our algorithm, iLA-SDP based on BM (i.e., iLA-SDP(BM) in the figure), exhibits nearly linear time complexity. Here, iLA-SDP refers to the original method for solving SDP, where the time cost grows with order $O(n^{3.5})$. Figure~\ref{fig:logtime} reveals that the iLA-SDP(BM) curve tends to be parallel to that of $K$-means, indicating its nearly \emph{linear time complexity}. It is noticeable that iLA-SDP(BM) has a relatively large and constant initialization complexity, which dominates the time cost when is relatively small. The iterative method of BM, which involves optimizing multiple parameters and selecting a low rank representation (we choose), may also contribute to this constant initialization cost. Nonetheless, our algorithm demonstrates superior performance over the original iLA-SDP, which has a super linear time complexity.

The space complexity of iLA-SDP (solved using BM) is of order $O(n)$, which is comparable to EM and KM. The space complexity of iLA-SDP comprises two parts: solving the SDP using BM, which has a space complexity of $O(n)$ , and solving the covariance matrices using the assignment matrices. The assignment matrices ($\bR^{n\times n}$) have low-rank representations ($\bR^{n\times r}$), resulting in a space complexity of $O(n)$ as well.

\begin{figure}[h!] 
   \centering
   \includegraphics[trim={4.5cm 10cm 4.5cm 9cm},clip,scale=0.8]{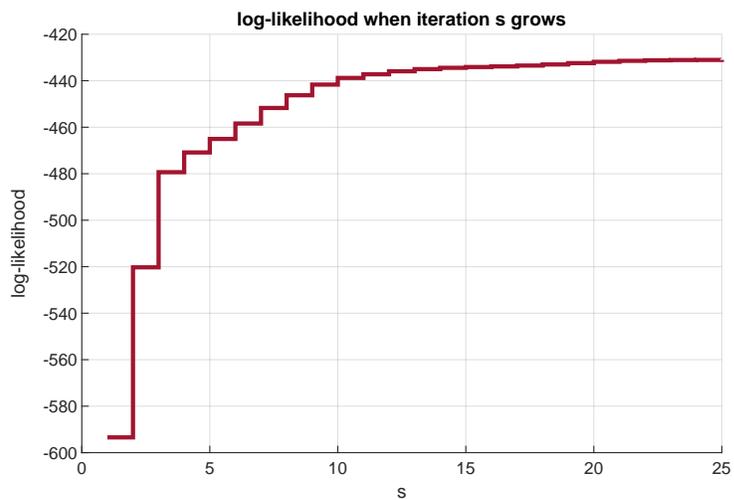}
   \caption{Log-likelihood (up to some constant) as iteration $s$ grows for LA-SDP. }
   \label{fig:loglik}
\end{figure}

\subsection{More discussions about iLA-SDP}\label{app:dis}

iLA-SDP is motivated by a more realistic scenario where clusters have heterogeneous or unknown covariance matrices. If the data structure is well-captured by simpler models such as SDP with isotropic or diagonal covariance matrices, then we can tailor the iLA-SDP covariance matrix updating procedure by incorporating this information (e.g., adding a penalty term on off-diagonal entries in high-dimensional cases). 

On the other hand, EM algorithm is known to be sensitive to initialization (in particular when the initialization uncertainty is large). As a result, we prefer to use iLA-SDP over EM because it is more stable in the face of large fluctuations in initialization, as demonstrated in Figure~\ref{fig:error_rates} and Figure~\ref{fig:barplot}. However, iLA-SDP may not be significantly better than the original SDP in cases where the data is high-dimensional and heterogeneous, as high-dimensional covariance matrix estimation is often inaccurate. This issue also affects the EM algorithm. To mitigate this problem, one can use PCA to perform dimension reduction, which projects the data onto several leading eigenspaces that capture most of the data variation. This pre-processing step enables us to extract the heterogeneous covariance matrix structure and improve the clustering performance.

Furthermore, our experiment \emph{Failure of EM vs SDP.} in Section~\ref{subsec:connections} shows the superior performance of LA-SDP in a scenario where two clusters are located close to each other, while the third cluster is far away from them. This situation is common in practical cases involving more than three clusters, where some clusters may share common structures and are therefore relatively close together, and some other clusters are far away. In such cases, we expect LA-SDP to perform better than the EM algorithm, as our method can handle these configurations with hierarchical structures.

\subsection{Real data application}\label{appendix:app}
For the handwriting digits dataset and CIFAR-10 dataset, we get the training set and test set directly from the original source and then applied dimension reduction on the training sets to get the low dimension representations. After that, we applied different methods on the test sets. To add randomness to the initialization, we randomly select a subset of the initialization for each repetition with $(1-\alpha)$ proportion, where we choose $\alpha =0.05.$ Similar for the landsat satellite dataset, the training and test sets are obtained from UCI machine learning repository and we ensure the randomness only comes from the initialization for KM, iLA-SDP and EM.

\subsection{Proof of the theorems and propositions}\label{appendix:proof}
In this section, we provide the proofs for the Proposition~\ref{prop:asdp_to_sdp}, Proposition~\ref{prop:soft_lasdp} and a sketch proof of Theorem~\ref{thm:main_thm}. 
The proof of the main theorem follows the track from the paper solving the exact recovery for original SDP~\citep{chen2021cutoff} and we will show the main differences in our proof.

First, we provide explicit expressions of some constants appearing in  Theorem~\ref{thm:main_thm} below:

\[
E_1=\frac{4(1+2\delta)M^{5/2}}{(1-\beta)^2\eta^2 } \left(M+\sqrt{M^2+ \frac{(1-\beta)^2}{(1+\delta)}\frac{p}{m\log n}+C_4 R_n} \right)
\]
with 
\[
R_n=\frac{(1-\beta)^2}{(1+\delta)\log n}\left( \frac{\sqrt{p\log n}}{\underline{n}}+\frac{\log n}{ \underline{n}}\right),
\]
and
\begin{equation}
\begin{aligned}
\label{equ:e2}
&E_2 = \frac{C_5(M-1)^3M^2 }{(1-\beta)(1-\eta) } \left(\frac{p}{\log n}+1\right)+ \frac{C_6 K^2(1-\beta)}{\beta}\\
&\cdot\min\left\{\frac{1  }{\beta(M-1)^2 }\frac{n}{m}\left(1+\frac{\log p}{\log n}\right) \frac{p}{\log n},\; \frac{(M-1)M^2}{\beta}\left(\sqrt{\frac{p^3}{\log n}}+\sqrt{p\log n} \right)\frac{n}{\sqrt{m}}\right\}.
\end{aligned}
\end{equation}

\subsubsection{Proof of Proposition~\ref{prop:asdp_to_sdp}}
\emph{Proposition~\ref{prop:asdp_to_sdp}} ({\bf SDP relaxation for $K$-means is a special case of LA-SDP}). 
Suppose $\Sigma_k = \sigma^2 \Id_p$ for all $k\in[K]$. Let $\hat{Z}$ be the solution to~(\ref{eqn:kmeans_sdp}) that achieves maximum $M_1$ and $\hat{Z}_k,k=1,\dots,K,$ be the solution to~(\ref{eqn:kmeans_sdp}) with maximum $M_2$. Then $M_1=M_2$. And $\hat{Z}=\sum_{k=1}^K \hat{Z}_k$, if $\hat{Z}$ is unique in~(\ref{eqn:kmeans_sdp}).

\emph{Proof of Proposition~\ref{prop:asdp_to_sdp}} If $\Sigma_k=\sigma^2\Id_p,\; \forall k\in[K].$ Then from (\ref{eqn:A_k}) we have 
$$A_k\equiv\frac{1}{2}\left[\diag(X^T X)\vone_{n}^T+\vone_{n}\diag(X^T  X)^T\right]+ X^T X,\; \forall k\in[K].$$
This implies that (\ref{eqn:kmeans_asdp}) can be written as
\begin{equation}
\label{eqn:kmeans_asdp_re}
\begin{gathered}
\hat{Z}_1,\dots,\hat{Z}_K = \arg\max_{Z_1,\dots Z_K \in \bR^{n \times n}}  \Big\langle X^T X , \big(\sum_{k=1}^K Z_k\big) \Big\rangle\\
\mbox{subject to } Z_k \succeq 0, \; \tr\big(\sum_{k=1}^K Z_k\big) = K, \; \big(\sum_{k=1}^K Z_k\big) \vone_{n} = \vone_{n}, \; Z_k \geq 0, \; \forall \; k\in [K],
\end{gathered}
\end{equation}
Since $\Big\langle \diag(X^T X)\vone_{n}^T, \big(\sum_{k=1}^K Z_k\big) \Big\rangle=\tr(X^TX)$, which is a constant in the optimization problem (\ref{eqn:kmeans_asdp_re}).  Now suppose $\hat{Z}$ is a solution to (\ref{eqn:kmeans_sdp}) that achieves maximum $M_1$ and $\hat{Z}_k,k=1,\dots,K,$ is the solution to~(\ref{eqn:kmeans_asdp_re}) that achieves maximum $M_2$, then we have
\[
 \Big\langle X^T X , \big(\sum_{k=1}^K Z_k\big) \Big\rangle\leq M1,
\]
\[
 \Big\langle X^T X , \big(\sum_{k=1}^K \tilde{Z}_k\big) \Big\rangle\leq M2,
\]
where $\tilde{Z}_1:=\hat{Z},\; \tilde{Z}_2=\cdots=\tilde{Z}_K=0.$ In other words, $M_1=M_2,$ which finishes the proof. If $\hat{Z}$ is unique in~(\ref{eqn:kmeans_sdp}), then we have $\hat{Z}=\sum_{k=1}^K \hat{Z}_k$ since both of them achieve the maximum in~(\ref{eqn:kmeans_sdp}). \qed

\subsubsection{Proof of Proposition~\ref{prop:soft_lasdp}}
\emph{Proposition~\ref{prop:soft_lasdp}} ({\bf iLA-SDP is a soft clustering method}). 
If $\rank(Z_k)=1,$ then there exists weights $(w_{k,1},\dots,w_{k,n})$ such that $\hat{\Sigma}_k$ in Lemma~\ref{lem:assign_to_cov} can be written as
\begin{equation}
    \label{eqn:LA-SDP_updating_rule}
    \hat{\Sigma}_k:=\frac{1}{n_k}\sum_{i=1}^n w_{k,i} (X_i-\hat{\mu}_k)(X_i-\hat{\mu}_k)^\top \quad \text{with} \quad \hat{\mu}_k:=\frac{1}{n_k} \sum_{i=1}^n w_{k,i}X_i,
\end{equation}
where $n_k = \sum_{i=1}^n w_{k,i}$.

\emph{Proof of Proposition~\ref{prop:soft_lasdp}} If $Z_k$ is rank $1,$ then there exists $a\in \bR^n$ such that $Z_k=aa^T.$ Let $w_{k}:=a^T\vone\cdot a,$ then we have 
\[
Z_k=\frac{w_{k}w_{k}^T}{w_{k}^T\vone},
\]
i.e., $Z_{k,ij}=\frac{w_{k,i}w_{k,j}}{\sum_{i=1}^nw_{k,i}}.$ Finally, by plugging in the expression of $Z_{k,ij}$ with $w_{k,i}$ we can get the target expression for $\hat{\Sigma}_k.$\qed

\subsubsection{Sketch proof of Theorem~\ref{thm:main_thm}}
\emph{Theorem~\ref{thm:main_thm}} ({\bf Exact recovery for LA-SDP}). 
Suppose there exist constants $\delta >0$ and $\beta\in(0,1)$ such that
\[
\log n \geq \max\left\{\frac{(1-\beta)^2}{\beta^2}, \frac{(1-\beta)(1-\eta)K^2}{\beta^2 \max\{(M-1)^2,1\}}  \right\}\frac{ C_1 n}{ m},\; \delta\leq  \frac{\beta^2}{(1-\beta)^2}\frac{ C_2 M^{1/2}}{ K},\; m\geq\frac{4(1+\delta)^2}{\delta^2}.
\]
If
\begin{equation}
\begin{aligned}
\label{eqn:lower_bound_SNR}
\Delta^2\geq (E_1+E_2)\log n,\;\text{and}\;\min_{k\ne l} D_{(k,l)}\geq C_3(1+\log n/p+p/n),
\end{aligned}
\end{equation}
where
\[
E_1=\frac{4(1+2\delta)M^{5/2}}{(1-\beta)^2\eta^2 } \left(M+\sqrt{M^2+ \frac{(1-\beta)^2}{(1+\delta)}\frac{p}{m\log n}+C_4 R_n} \right)
\]
with 
\[
R_n=\frac{(1-\beta)^2}{(1+\delta)\log n}\left( \frac{\sqrt{p\log n}}{\underline{n}}+\frac{\log n}{ \underline{n}}\right),
\]
and
\begin{equation}
\begin{aligned}
\label{equ:e2}
&E_2 = \frac{C_5(M-1)^3M^2 }{(1-\beta)(1-\eta) } \left(\frac{p}{\log n}+1\right)+ \frac{C_6 K^2(1-\beta)}{\beta}\\
&\cdot\min\left\{\frac{1  }{\beta(M-1)^2 }\frac{n}{m}\left(1+\frac{\log p}{\log n}\right) \frac{p}{\log n},\; \frac{(M-1)M^2}{\beta}\left(\sqrt{\frac{p^3}{\log n}}+\sqrt{p\log n} \right)\frac{n}{\sqrt{m}}\right\};
\end{aligned}
\end{equation}
then the LA-SDP achieves exact recovery, or $\hat Z=Z^*$, with probability at least $1-C_7 K^3 n^{-\delta}$ for some universal constants $C_1,\dots,C_7$.  

\emph{Sketch of the proof.}
Recall that we let $G_1^*, \dots, G_K^*$ be the true partition of the index set $[n] := \{1, \dots, n\}$ such that if $i \in G_k^*$, then
\begin{equation}
    \label{eqn:gmm}
    X_i=\mu_k+\epsilon_i,
\end{equation} where $\mu_k \in \bR^p$ is the true center of the $k$-th cluster $G_k^*$ ($G_k$ for simplicity) and $\epsilon_i$ is an i.i.d. random Gaussian noise $N(0,\Sigma_k)$. First we can write down the dual problem:
\[
\min_{\substack{\lambda\in\bR,\alpha\in\bR^n,\\ B_k\in\bR^{n\times n}}} \lambda K+\alpha^T\vone_n,\; \text{subject to} \; B_k\geq0, \; \lambda \Id_n+\frac{1}{2}(\alpha \vone_n^T+\vone_n \alpha^T)-A_k-B_k\succeq 0,\; \forall k\in[K].
\]
Denote $Z_k^\ast:= \frac{1}{|G_k|}\vone_{G_k}\vone_{G_k}^T,\; \forall k\in[K]$ then it can be shown that the sufficient conditions for the solution of SDP to be $Z_k=Z_k^\ast,\; \forall k\in[K]$ are
\[
B_k\geq 0; \tag{C1}
\]
\[
W_k:=\lambda \Id_n+\frac{1}{2}(\alpha \vone_n^T+\vone_n \alpha^T)-A_k-B_k\succeq 0; \tag{C2}
\]
\[
\tr(W_kZ_k^\ast)=0; \tag{C3}
\]
\[
\tr(B_kZ_k^\ast)=0. \tag{C4}
\]

It can be verified that if we can find symmetric $B_k$ such that
\[
B_{k,G_kG_k}=0;
\]
\begin{align*}
[B_{k,G_lG_k} \vone_{G_k}]_i&=-\frac{n_k+n_l}{2n_l}\cdot \lambda\\
&+\frac{n_k}{2}[(\|\Sigma_k^{-1/2}(\bar{X_k}-X_i)\|^2+\log|\Sigma_k|)-(\|\Sigma_l^{-1/2}(\bar{X_l}-X_i)\|^2+\log|\Sigma_l|)];\\
[B_{k,G_lG_l} \vone_{G_l}]_j&=[A_{l,G_{l}G_l} \vone_{G_l}]_j-[A_{k,G_{l}G_l} \vone_{G_l}]_j;\\
[B_{k,G_{l'}G_l} \vone_{G_l}]_j&=[B_{l,G_{l'}G_l} \vone_{G_l}]_j+[A_{l,G_{l'}G_l} \vone_{G_l}]_j-[A_{k,G_{l'}G_l} \vone_{G_l}]_j,
\end{align*}

for any triple pairs $(k,l,l')$ that are mutually distinct and $i\in G_k, \; j\in G_l$. Then (C3) and (C4) hold. In fact, the target matrices can be defined through
\begin{equation}
   B^\#_{k,G_{l'}G_l}:=\frac{B_{k,G_{l'}G_l} \vone_{G_l}\vone_{G_{l'}}^TB_{k,G_{l'}G_l} }{\vone_{G_{l'}}^T B_{k,G_{l'}G_l} \vone_{G_l}}, 
\end{equation}

for any $k\in[K],\; (l',l)\ne(k,k)$. Furthermore, the construction of $B_k$ shows that $B_k\vone_{G_{l}}=0, \; \forall (k,l)$ pairs.

The following two lemma gives the sufficient conditions for (C1).
\begin{lem}[\bf Separation bound on the covariance matrices]
\label{lem:sepa2}
Let $\lambda_1,\dots,\lambda_p$ correspond to the eigenvalues of $(\Sigma_l^{1/2}\Sigma_k^{-1}\Sigma_l^{1/2}-\Id_p ) $ and define $D_{(k,l)}:= \frac{\sum_{i=1}^p\left(\lambda_i -\log(1+\lambda_i)\right)}{p\max_{i}|\lambda_i |}.$
If there exists constant $C$ such that
\[
\min_{k\ne l} D_{(k,l)}\geq C(1+\log n/p+p/n),
\]
then 
\[
\bP\Big([A_{l,G_{l}G_l} \vone_{G_l}]_j-[A_{k,G_{l}G_l} \vone_{G_l}]_j\geq 0,\; \text{for all }(k,l)\in[K]^2~\text{and}~j\in G_l\Big)\geq 1-CK^2/n.
\]
\end{lem}

\begin{lem}[\bf Separation bound on the centers]
\label{lem:sepa1}

Let $\delta>0,\;\beta\in(0,1),\; \eta\in(0,1)$. If we have
\[
\Delta^2\geq \frac{4(1+\delta)M^2}{(1-\beta)^2\eta^2}\left[M^{3/2}+\sqrt{M^3+\frac{(1-\beta)^2M}{(1+\delta)}\frac{p+2\sqrt{p\log(nK)}+4\log(nK)}{m\log n}}\right]\log n,
\]
and
\begin{align*}
&\Delta^2\geq \frac{M^2(M-1)^2}{(1-\beta)^2(1-\eta)^2}\cdot\\
&\left(1+\frac{2(1-\beta)(1-\eta)}{M}[3\log M+4M(M-1)(p+2\sqrt{p\log(nK)}+4\log(nK))]\right),
\end{align*}
then
\begin{align*}
 &\bP\Big(\|\Sigma_l^{-1/2}(\bar{X_l}-X_j)\|^2+\log|\Sigma_l|)-(\|\Sigma_{l'}^{-1/2}(\bar{X_{l'}}-X_j)\|^2+\log|\Sigma_{l'}|)\\
 &-\frac{2}{n_l}\big|[A_{l,G_{l'}G_l }\vone_{G_l}]_j-[A_{k,G_{l'}G_l }\vone_{G_l}]_j\big| \geq \frac{\beta}{M} \|\Sigma_l^{-1/2}(\mu_l-\mu_{l'}) \|^2+(n_l^{-1}+n_{l'}^{-1})p-r_{k,l,l'}, \\
&\text{for all triple}~(k,l,l')\in[K]^3
~\text{with}~(k,l,l')\ne (k,k,k)~\text{and}~j\in G_{l'}\Big)  \\
& \leq \frac{CK^3}{n^\delta},
\end{align*}
where
\[
r_{k,l,l'}=4\sqrt{\frac{\log(nK)}{n_l}} \|\Sigma_l^{-1/2}(\mu_l-\mu_{l'}) \|+2(n_l^{-1}+n_{l'}^{-1})\sqrt{2p\log(nK)}+4n_{l'}^{-1}\log(nK).
\]
for some large constant $C$.
\end{lem}
The proof of Lemma~\ref{lem:sepa1} follows the similar steps from the original paper~\citep{chen2021cutoff}. The two lemmas imply that (C1) can hold with high probability if the separation condition in the assumption holds. The remaining part is to verify the (C2).

Denote $\Gamma = \text{span}\{\vone_{G_k}:k\in[K]\}^\perp$ be the othogonal complement of the linear space spanned by $\vone_{G_k},\; k\in[K].$ Note that $W_k\vone_{G_l}=0,\; \forall (k,l)\in[K]^2,$ we only need to check for $v\in \Gamma,$ 
\[
v^TW_kv\ge0, \;\forall k\in[K].
\]
Note that $v^T\vone_{G_k}=0,$ we have 
\[
v^TW_kv=\lambda \|v\|^2-S_k(v)-T_k(v),
\]
where $S_k(v):=v^T A_k v=v^TX^T\Sigma_k^{-1}Xv,$ and $T_k(v)=v^TBv.$ By concentration bound we can get 
\[
\bP(S_k(v)\leq MK(\sqrt{n}+\sqrt{p}+\sqrt{2\log n}),\; \text{for all}~k\in[K])\geq 1-\frac{K}{n}.
\]
For $T_k(v)$, first we define
\[
V_{k,ll'}^{(1)}:=\langle \Sigma_{l'}^{1/2}\Sigma_{l}^{-1}(\mu_{l'}-\mu_l),\sum_{j\in G_{l'}}v_j\epsilon_j \rangle;
\]
\[
V_{k,ll'}^{(2)}:= \langle \bar{\epsilon}_{l'}-\Sigma_{l'}^{1/2}\Sigma_{l}^{-1/2}\bar{\epsilon}_{l}, \sum_{j\in G_{l'}}v_j\epsilon_j \rangle;
\]
\[
V_{k,ll'}^{(3)}:= \frac{1}{2}\sum_{j\in G_{l'} } \epsilon_j^T \Sigma_{l'}^{1/2}(\Sigma_{l}^{-1}-\Sigma_{l'}^{-1})\Sigma_{l'}^{1/2}\epsilon_jv_j ;
\]
\[
V_{k,ll'}^{(4)}:= \frac{1}{n_l}\sum_{j\in G_{l'} } ([A_{l,G_{l'}G_l }\vone_{G_l}]_j-[A_{k,G_{l'}G_l }\vone_{G_l}]_j)v_j\cdot \vone(l\ne l').
\]
Then we can write $T_k(v)$ as
\[
T_k(v):=\sum_{l\ne l'}\frac{n_ln_{l'}}{\vone_n^T B_k \vone_n}(T_{k,ll'}^{(1)}+T_{k,ll'}^{(2)}+T_{k,ll'}^{(3)}+T_{k,ll'}^{(4)}+T_{k,ll'}^{(5)}),
\]
where
\[
T_{k,ll'}^{(1)}:=V_{k,ll'}^{(1)}\cdot V_{k,l'l}^{(1)};
\]
\[
T_{k,ll'}^{(2)}:=V_{k,ll'}^{(2)}\cdot V_{k,l'l}^{(2)};
\]
\[
T_{k,ll'}^{(3)}:=V_{k,ll'}^{(1)}\cdot V_{k,l'l}^{(2)}+V_{k,ll'}^{(2)}\cdot V_{k,l'l}^{(1)};
\]
\[
T_{k,ll'}^{(4)}:=(V_{k,ll'}^{(3)}+V_{k,ll'}^{(4)})\cdot (V_{k,l'l}^{(1)}+V_{k,l'l}^{(2)})+(V_{k,ll'}^{(1)}+V_{k,ll'}^{(2)})\cdot (V_{k,l'l}^{(3)}+V_{k,l'l}^{(4)});
\]
\[
T_{k,ll'}^{(5)}:=(V_{k,ll'}^{(3)}+V_{k,ll'}^{(4)})\cdot  (V_{k,l'l}^{(3)}+V_{k,l'l}^{(4)}).
\]
Now we choose $\lambda=p+\frac{\beta}{4M}m\Delta^2,$ which implies that 
\[
\vone_n^T B_k \vone_n\geq \frac{n_ln_{l'}}{8}\frac{\beta}{M}\max\{\|\Sigma_{l'}^{-1/2}(\mu_l-\mu_{l'})\|^2,\|\Sigma_{l}^{-1/2}(\mu_l-\mu_{l'})\|^2 \}.
\]
From concentration bounds for Gaussians we have for all triple $(k,l,l')\in[K]^3$ such that $(k,l,l')\ne(k,k,k)$,
\[
\left|\sum_{l\ne l'}\frac{n_ln_{l'}}{\vone_n^T B_k \vone_n}T_{k,ll'}^{(1)}\right|\leq \frac{CM^2}{\beta}\cdot (n+\sqrt{2nK\log n}+2K\log n) \|v\|^2;
\]
\[
\left|\sum_{l\ne l'}\frac{n_ln_{l'}}{\vone_n^T B_k \vone_n}T_{k,ll'}^{(2)}\right|\leq \frac{CM^3}{1-\beta}\cdot (\delta \sqrt{mp\log n}+\sqrt{mp\log^7 n}/\underline{n}) \|v\|^2;
\]
\[
\left|\sum_{l\ne l'}\frac{n_ln_{l'}}{\vone_n^T B_k \vone_n}T_{k,ll'}^{(5)}\right|\leq \frac{CK^2}{\beta}\cdot (\frac{1-\beta}{(M-1)^2M}(p+pM\log p/\log n)+M(M-1))n \|v\|^2,
\]
Or
\[
\left|\sum_{l\ne l'}\frac{n_ln_{l'}}{\vone_n^T B_k \vone_n}T_{k,ll'}^{(5)}\right|\leq \frac{CK^2M(1-\beta)(M-1)}{\beta}\cdot (\sqrt{\frac{p^3m}{\log n}}+\sqrt{pm\log n })n \|v\|^2,
\]
with probability $\geq 1-CK^3/n^\delta$ for some constant $C$.

Note that by assumption we have $\Delta^2\geq \frac{C(M-1)^3M^2}{(1-\beta)(1-\eta)}(p+\log n)+\frac{CM^3}{(1-\beta)}\sqrt{(1+\delta)p\log n/m}$ and the fact that the remaining terms of $T_{k,ll'}$ can be bounded by the above inequalities up to multiplied by some constant, we can directly verify that (C2) is true under our assumptions.\qed

\emph{Lemma~\ref{lem:sepa2}} ({\bf Separation bound on the covariance matrices}).
Let $\lambda_1,\dots,\lambda_p$ correspond to the eigenvalues of $(\Sigma_l^{1/2}\Sigma_k^{-1}\Sigma_l^{1/2}-\Id_p ) $ and define $D_{(k,l)}:= \frac{\sum_{i=1}^p\left(\lambda_i -\log(1+\lambda_i)\right)}{p\max_{i}|\lambda_i |}.$
If there exists constant $C$ such that
\[
\min_{k\ne l} D_{(k,l)}\geq C(1+\log n/p+p/n),
\]
then 
\[
\bP\Big([A_{l,G_{l}G_l} \vone_{G_l}]_j-[A_{k,G_{l}G_l} \vone_{G_l}]_j\ge0,\; \text{for all }(k,l)\in[K]^2~\text{and}~j\in G_l\Big)\geq 1-CK^2/n.
\]

\emph{Sketch of the proof.}
Let $T:=[A_{l,G_{l}G_l} \vone_{G_l}]_j-[A_{k,G_{l}G_l} \vone_{G_l}]_j,\; B:=\Sigma_l^{1/2}\Sigma_k^{-1}\Sigma_l^{1/2}-\Id_p$ then by definition we have
\begin{align*}
 T&=-\sum_{i=1}^p\log(\lambda_i+1)+\sum_{i=1}^p \lambda_i\\
 &+\frac{1}{2}\langle B,\epsilon_j \epsilon_j^T-\Id_p \rangle\\
 &-\frac{1}{2}\langle B,\frac{1}{n_l}\sum_{t\in G_l}\epsilon_t \epsilon_j^T+\epsilon_j\Big(\frac{1}{n_l}\sum_{t\in G_l}\epsilon_t\Big)^T\rangle\\
 &+\frac{1}{2}\langle B,\frac{1}{n_l}\sum_{t\in G_l}\epsilon_t\epsilon_t^T -\Id_p \rangle,
\end{align*}
where the last three terms can be bounded by concentration bounds for Gaussians. \qed

\end{document}